\documentclass[sigconf]{acmart}

\usepackage{mathrsfs}
\usepackage{color}
\usepackage{booktabs}
\usepackage{multirow}
\usepackage{indentfirst}
\usepackage{bm}
\usepackage{makecell}
\usepackage[T1]{fontenc}
\usepackage{subfigure}
\usepackage{url}

\AtBeginDocument{%
  \providecommand\BibTeX{{%
    \normalfont B\kern-0.5em{\scshape i\kern-0.25em b}\kern-0.8em\TeX}}}

\copyrightyear{2020}
\acmYear{2020}
\setcopyright{acmcopyright}\acmConference[MM '20]{Proceedings of the 28th ACM International Conference on Multimedia}{October 12--16, 2020}{Seattle, WA, USA}
\acmBooktitle{Proceedings of the 28th ACM International Conference on Multimedia (MM '20), October 12--16, 2020, Seattle, WA, USA}
\acmPrice{15.00}
\acmDOI{10.1145/3394171.3413618}
\acmISBN{978-1-4503-7988-5/20/10}

\begin{document}
	\fancyhead{}
	
	\title{Modeling Temporal Concept Receptive Field Dynamically\\ for Untrimmed Video Analysis}


\author{Zhaobo Qi$^{1,2}$, Shuhui Wang$^{2, *}$, Chi Su$^{3}$, Li Su$^{1, *}$, Weigang Zhang$^{4}$, Qingming Huang$^{1,2,5}$}
\thanks{$^\ast$Corresponding author.}
\affiliation{%
	\institution{$^1$ University of Chinese Academy of Sciences, Beijing, China}
	\institution{$^2$ Key Lab of Intell. Info. Process., Inst. of Comput. Tech., CAS, Beijing, China}
	\institution{$^3$ Kingsoft Cloud, Beijing, China}
	\institution{$^4$ Harbin Inst. of Tech, Weihai, China}
	\institution{$^5$ Peng Cheng Laboratory, Shenzhen, China}
}
\email{zhaobo.qi@vipl.ict.ac.cn, wangshuhui@ict.ac.cn, suchi@kingsoft.com, suli@ucas.ac.cn,wgzhang@hit.edu.cn, qmhuang@ucas.ac.cn}

%

\renewcommand{\shortauthors}{Trovato and Tobin, et al.}

\begin{abstract}
Event analysis in untrimmed videos has attracted increasing attention due to the application of cutting-edge techniques such as CNN. As a well studied property for CNN-based models, the receptive field is a measurement for measuring the spatial range covered by a single feature response, which is crucial in improving the image categorization accuracy. In video domain, video event semantics are actually described by complex interaction among different concepts, while their behaviors vary drastically from one video to another, leading to the difficulty in concept-based analytics for accurate event categorization. To model the concept behavior, we study temporal concept receptive field of concept-based event representation, which encodes the temporal occurrence pattern of different mid-level concepts. Accordingly, we introduce temporal dynamic convolution~(TDC) to give stronger flexibility to concept-based event analytics. TDC can adjust the temporal concept receptive field size dynamically according to different inputs. Notably, a set of coefficients are learned to fuse the results of multiple convolutions with different kernel widths that provide various temporal concept receptive field sizes. Different coefficients can generate appropriate and accurate temporal concept receptive field size according to input videos and highlight crucial concepts. Based on TDC, we propose the temporal dynamic concept modeling network~(TDCMN) to learn an accurate and complete concept representation for efficient untrimmed video analysis. Experiment results on FCVID and ActivityNet show that TDCMN demonstrates adaptive event recognition ability conditioned on different inputs, and improve the event recognition performance of Concept-based methods by a large margin. Code is available at ~\url{https://github.com/qzhb/TDCMN}.
\end{abstract}



\maketitle

\section{Introduction}
\label{introduction}

The receptive field is defined as the region in the input space that a particular CNN feature is looking at \cite{receptive-field-definition}. It has been studied extensively to assist us to construct more expressive neural networks in recent years for image recognition~\cite{szegedy2016rethinking}, semantic segmentation~\cite{Yu2016Multi}, object detection~\cite{li2019scale-aware}, etc. Most methods have been developed to deal with the spatial feature for image-based tasks. However, in video domain, this issue has been less studied in literature. 

On the other hand, video analysis has attracted considerable attention due to its extensive applications. Neural-network-based methods have been proposed to promote the development of this field, such as \cite{karpathy2014large, feichtenhofer2016convolutional, wang2016temporal, tran2015learning, carreira2017quo, feichtenhofer2019slowfast}. These methods extract feature representations directly from frames or optical flow, which is recognized as Appearance-based methods. The construction of these models inherits from image-based analysis models and considers the characteristics of video data itself.

\begin{figure}[t]
	\centering
	\subfigure[Simple event categories. Small temporal receptive field is needed.]{
		\includegraphics[scale=0.4]{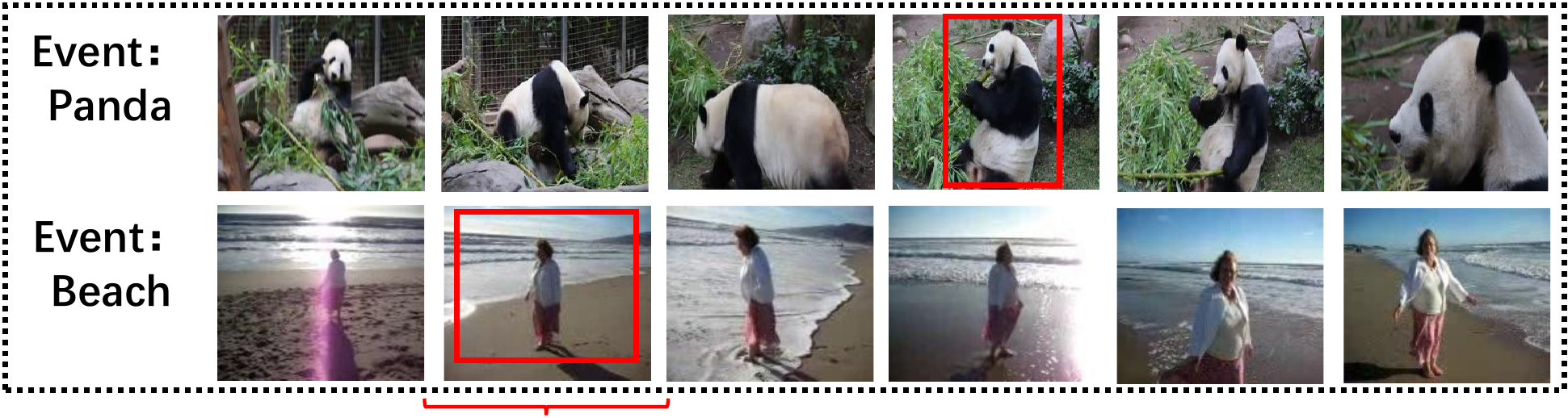}
		\label{fig:motivation-case.1}
	}
	\quad
	\subfigure[Subtle event categories. Large temporal receptive field is needed.]{
		\includegraphics[scale=0.4]{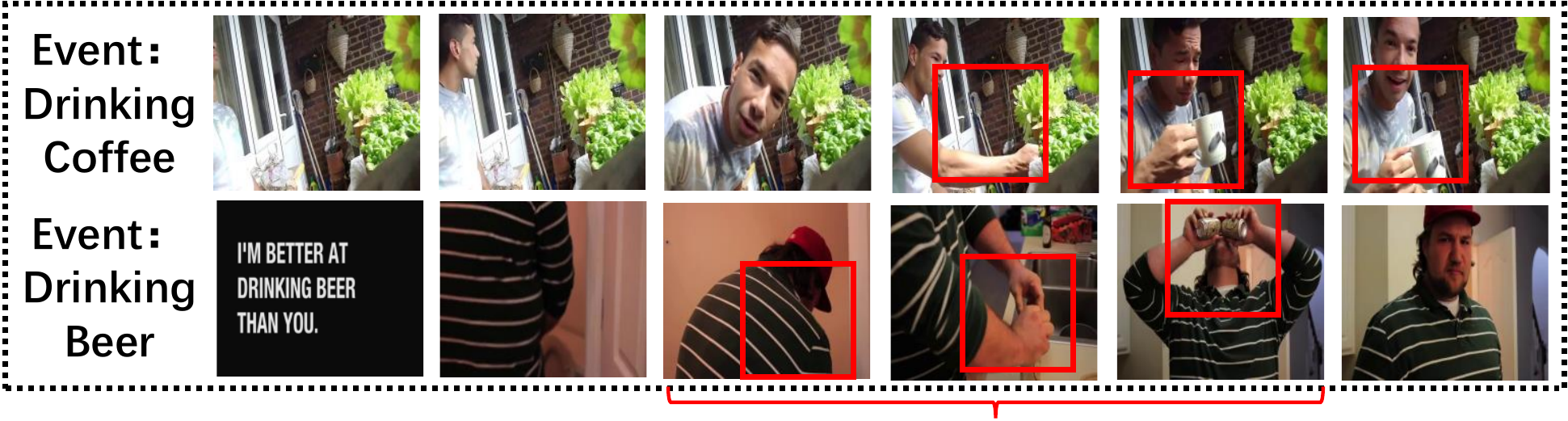}
		\label{fig:motivation-case.2}
	}
	\quad
	\subfigure[Large intra-class variations. Various temporal receptive fields are needed.]{
		\includegraphics[scale=0.4]{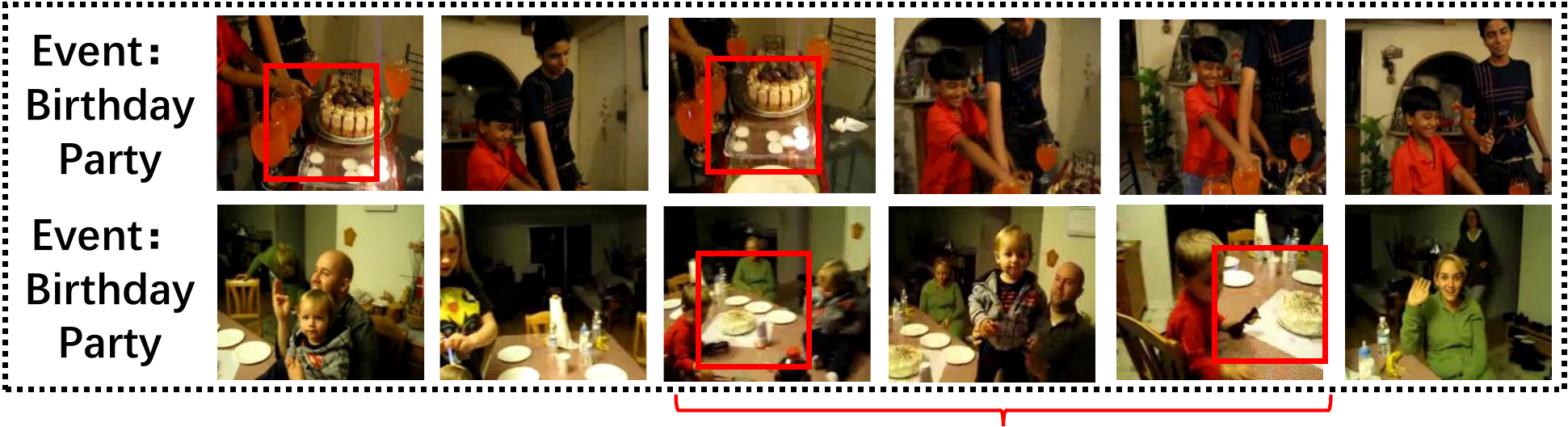}
		\label{fig:motivation-case.3}
	}
	\quad
	\centering
	\caption{ The difficulty of making recognition decisions relates to the event category to be classified. For different cases, the needed temporal concept receptive field size for capturing crucial concepts~(marked as red bounding box) and differentiating them are varied.}
	\label{fig:motivation-case}
\end{figure}

Moreover, in order to produce more interpretable recognition results, some researches have focused on Concept-based event recognition methods~\cite{bhattacharya2014recognition, chang2015searching, chang2016they, chang2017semantic, fan2017complex, wu2016harnessing, xu2014event, yan2015complex}. These models harness simple aggregation method to get video-level concept representations. They ignore the temporal characteristics of concept representations, which limits the feature representation ability. Essentially, the temporal concept receptive field has not been fully exploited, which results in a significant performance gap between Concept-based and Appearance-based event recognition methods. 

The various temporal concept receptive field sizes related to the event categories to be classified is another key issue. For some simple event categories~(e.g., `panda' and `beach', as shown in Figure~\ref{fig:motivation-case.1}), crucial concepts existing in short time duration might be sufficient to recognize them. Therefore, the actually suitable temporal receptive field size is small. For some event categories that are subtle events~(e.g. `drinking coffee' and `drinking beer', as shown in Figure~\ref{fig:motivation-case.2}), concepts in a long time duration should be captured to differentiate them, which needs a large temporal receptive field size. This also holds for samples within the same event category due to large intra-event variations. For example, the concepts `cakes' useful for recognizing the event `birthday party'~(as shown in Figure~\ref{fig:motivation-case.3}) may occur at different time stamps, such as the beginning, the end, or throughout the video. Hence, multiple temporal receptive field sizes for different input videos tend to be more in demand. In a nutshell, a model that can adjust its temporal concept receptive field adaptatively based on different inputs will be more appropriate for flexible and accurate event recognition.

Considering the above-mentioned issues, we explore the temporal concept receptive field of concept-based event recognition models in this paper. As shown in Figure \ref{fig:framework-tdc}, we propose the temporal dynamic convolution~(TDC) to give stronger flexibility to concept-based event recognition methods. TDC can adjust the temporal concept receptive field size dynamically based on different inputs. The goal of TDC is to learn a set of coefficients to fuse the results of multiple convolutions with different kernel widths that can provide various temporal concept receptive field sizes. 
Based on the generated coefficients, we can construct appropriate and accurate temporal concept receptive field size for corresponding videos, which can assist us to highlight crucial concepts for efficient video recognition.

Based on the temporal dynamic convolution, we propose temporal dynamic concept modeling network~(TDCMN), a general Concept-based event recognition model augmented with the temporal dynamic convolution. Considering the existence of unique temporal pattern of each concept and the relationship between different types of concepts, our TDCMN contains two key elements, an intra-domain temporal dynamic concept modeling network~(InTDCM) and a cross-domain temporal dynamic concept modeling network~(CrTDCM). In InTDCMN, we apply TDC on each type of concept representation separately. In CrTDCMN, we construct two modeling pipelines, one of which is to apply TDC on the concatenation of all types of concept representations, and the other is to extend TDC to cross-domain temporal dynamic convolution. Our TDCMN can learn an accurate and complete concept representation for efficient untrimmed video analysis.

We conduct extensive experiments on two large-scale and challenging video datasets FCVID and ActivityNet. Experiment results show that TDCMN can achieve adaptive inference conditioned on input videos by adjusting its temporal concept receptive field size. Our TDCMN can improve the performance of Concept-based event recognition methods by a large margin. Besides, it can also obtain higher performance compared with some Appearance-based event recognition methods. 

The contributions of this paper are three-fold.
\begin{itemize}
	\item We propose to analysis video event from the temporal concept receptive field. Accordingly, we propose the dynamic temporal convolution, which can adjust temporal concept receptive field based on input adaptatively.
	\item  We propose the temporal dynamic concept modeling network~(TDCMN), a general concept-based event recognition model augmented with the temporal dynamic convolution. TDCMN can learn an accurate and complete concept representation for efficient untrimmed video analysis.
	\item TDCMN achieves adaptive inference conditioned on input videos by adjusting its temporal concept receptive field size. It can improve the performance of Concept-based methods by a large margin and obtain higher performance than some Appearance-based methods on two challenging datasets.
\end{itemize}

\section{Related Work}

We review related works from two aspects: video event recognition methods, dynamic receptive field. 

\subsection{Video Event Recognition Methods}

\textbf{Appearance-based methods.} With the rapid development of deep convolution neural networks, plenty of works have been proposed to learn vision and motion features via deep models for video event recognition, {\it e.g.}, 2D-CNN-based methods~\cite{karpathy2014large}, two-stream-based  methods~\cite{feichtenhofer2016convolutional, wang2016temporal} and 3D-CNN-based methods (C3D~\cite{tran2015learning}, I3D~\cite{carreira2017quo} and SlowFast~\cite{feichtenhofer2019slowfast}). These methods can achieve high performance, but lack of interpretability. Therefore, we focus on more interpretable event recognition models in this paper, and briefly review its recent process below. 


\textbf{Concept-based methods.} 
For concept-based methods, the basic is the selection of the concepts. Some define event-driven concepts~\cite{chen2014event,ye2015eventnet}, while others focus on using manually chosen concepts or concept libraries~\cite{bhattacharya2014recognition, chang2015searching, chang2016they,chang2017semantic, fan2017complex, wu2016harnessing, xu2014event, yan2015complex}. Ye {\it et al.}~\cite{ye2015eventnet} build a large scale event-specific concept library EventNet that covers as many real-world events and concepts as possible. In this paper, we consider three key elements when we select concepts, which can ensure the versatility and practicability of our method. First, the concept detectors can be obtained directly without training from scratch. Second, different types of concept detectors must be used. Third, the concept  categories of each type of concept detector must be varied and common. 

As for event analysis methods, Chang {\it et al.}~\cite{chang2017semantic} propose a semantic pooling approach which learns the relation between concepts and events through skip-gram model, and the concept relation is used to prioritize the video shot representations. Xu {\it et al.}~\cite{xu2014event} build up a multiple feature learning framework for complex event detection. Wu {\it et al.}~\cite{wu2016harnessing} introduce Object-Scene semantic Fusion (OSF) network for large-scale video understanding. Compared to these methods, we analysis concept representation from a new perspective, which is the temporal concept receptive field. We propose a temporal dynamic convolution~(TDC), which can adjust the temporal concept receptive field adaptively based on different input videos and guarantee more flexible and effective concept modeling.

\subsection{Dynamic Receptive Field}

The receptive field is defined as the region in the input space that a particular CNN feature is looking at (i.e. be affected by)\cite{receptive-field-definition}. It is crucial for constructing expressive or light-weight convolution neural networks.

Some prior works increase the receptive field of neural networks by constructing deeper models~\cite{he2016deep, szegedy2016inception-v4} with downsampling such as pooling operation or dilated convolution~\cite{Yu2016Multi}. These works can accumulate some fixed-sized receptive fields at the expense of high computational burden to increase the feature expression. Some prior works~\cite{de-brabandere2016dynamic, burkov2018deep, zhang2019accelerating, lioutas2020time-aware} focus on directly learning the convolution kernel parameters or the location offsets of convolution kernels. These works usually require a great number of parameters and are difficult to extend to multiple-layers neural networks. 

Some prior works~\cite{chen2019dynamic, li2019selective, yang2019condconv:} focus on predicting the coefficients to combine multiple static convolution kernels. These methods can make models more expressive or reduce redundant calculations in convolution neural networks. Especially, Chen {\it et al.}~\cite{chen2019dynamic} present a dynamic convolution to aggregate multiple parallel convolution kernels dynamically based on their attention, which can increase model expressive without increasing the network depth or width for constructing light-weight conventional neural networks. They use multiple convolution filters with the same kernel width. Instead, we utilize multiple convolution filters with different kernel widths in our work, which can provide multi-scale information. Li {\it et al.}~\cite{li2019selective} propose the SKNet to fuse the results of multiple convolution filters with different kernel sizes, which allows the neural network to adjust its receptive field size based input information adaptively. Inspired by this idea, we develop the dynamic convolution for video event analysis. In contrast to these prior methods, there are several differences. We utilize the dynamic convolution to analysis temporal information for video task. Instead, they use it to process spatial information for image-based tasks. Instead of injecting the dynamic convolution on all layers of deep neural networks, our proposed dynamic convolution is only utilized as a small but crucial part of our model, which can not increase significant computational burden. Moreover, in terms of calculating the fusion coefficients, we take into account the relationships within and between different convolution filters, which can make our model more flexible and expressive. Instead, prior works only consider the relation between different convolution filters.

%

\section{Proposed Methodology}
\label{proposed-method}

In this section, we will start with the general framework of concept-based event recognition model in~Section \ref{concept-based method}, which is the basic model of the following analysis. Then, we will introduce the temporal dynamic convolution~(TDC) in~Section \ref{tdc}. Finally, we propose the temporal dynamic concept modeling network~(TDCMN) in~Section \ref{tdcmn}.

\subsection{Concept-based Event Recognition Model}
\label{concept-based method}

The general framework of concept-based event recognition model is described below. First, each video $V$ is evenly divided into $N$ clips~$\left\{C_1, C_2, ..., C_N\right\}$ and $M$ frames are randomly sampled from each clip. Second, different types of concept detectors $\left\{D_1, D_2, ..., D_O\right\}$ are applied on all video clips to capture the initial concept representations. Specially, for each concept detector $D_{i}$, we feed all video clips into it and obtain the initial concept representation $\boldsymbol{X}_{i}=[x_{i,1}, x_{i,2}, ..., x_{i,N}]\in{\mathbb{R}^{L_i\times{N}}}$, where $L_i$ is the concept category number of $D_i$. We will show how to select and use these different types of concept detectors in detail in Section~\ref{subsection:choosen-detectors}. After that, the aggregation method such as max pooling is applied on $\boldsymbol{X}_{i}$ to obtain the aggregated concept representation. Next, we obtain the video level concept representation $X\in{\mathbb{R}^{1\times{L}}}$ by concatenating all types of aggregated concept representations, where $ L = \sum_{i=1}^O{L_i}$ is the total number of concept category of all concept detectors. Finally, a simple MLP network is used to predict the event categories. For the sake of simplicity, we use two types of concept detectors~$\left\{D_1, D_2\right\}$ and obtain two types of initial concept representations $\boldsymbol{X}_{1}\in{\mathbb{R}^{L_1\times{N}}}$ and $\boldsymbol{X}_{2}\in{\mathbb{R}^{L_2\times{N}}}$ in the following sections. 

\subsection{Temporal Dynamic Convolution}
\label{tdc}

\begin{figure}[t]
	\centering
	\subfigure[Results fusion module.]{
		\includegraphics[scale=0.5]{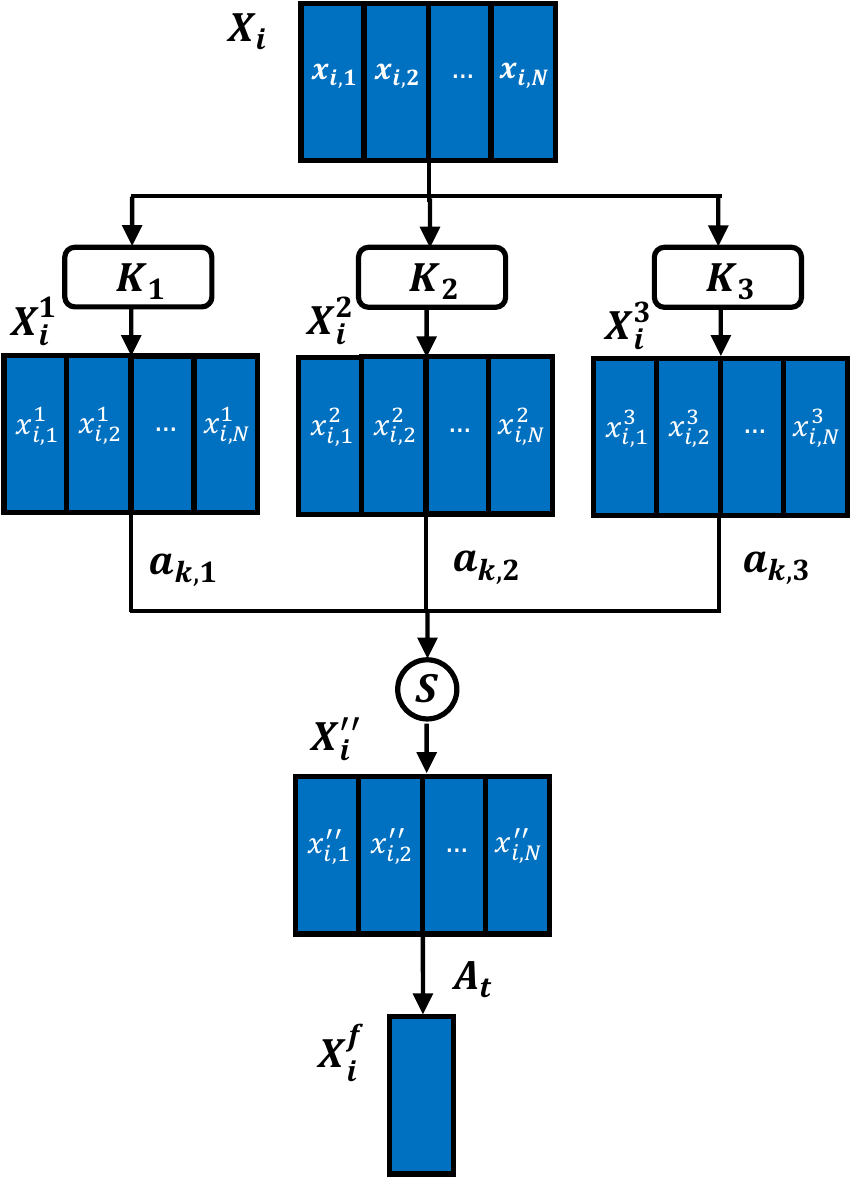}
		\label{fig:framework-tdcres}
	}
	\quad
	\hspace{-2ex}
	\subfigure[Coefficient generation module.]{
		\includegraphics[scale=0.55]{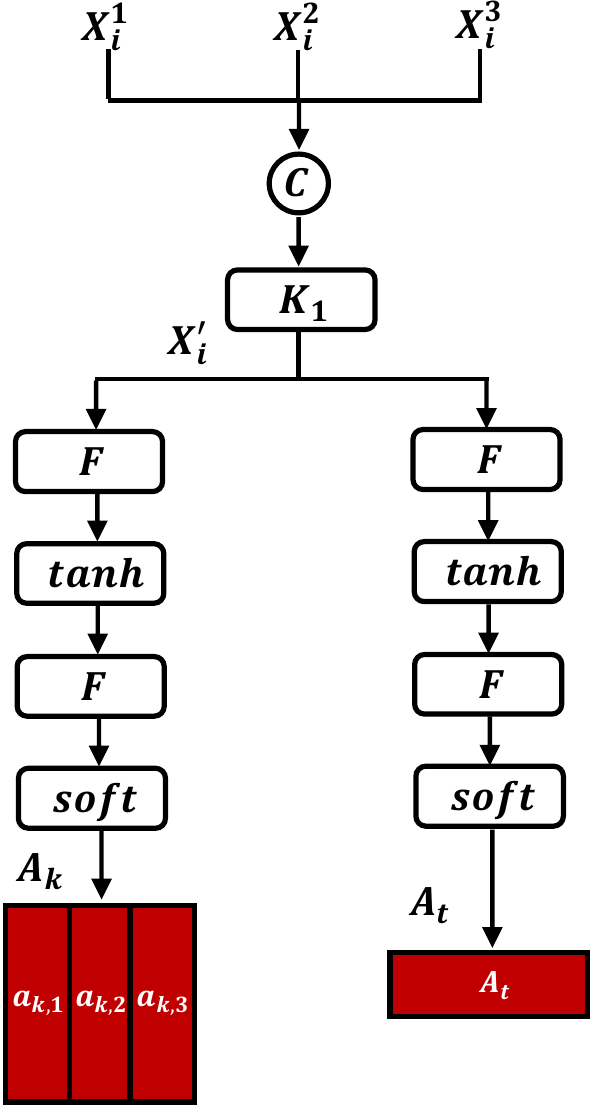}
		\label{fig:framework-tdccoe}
	}
	\quad
	\centering
	\caption{Temporal dynamic convolution. $K_{i}$ represents a 1d convolution operation. $S$, $C$, $F$, $tanh$, $soft$ represent summation operation, concatenation operation, fully connected layer, tanh function and softmax function, separately.}
	\label{fig:framework-tdc}
\end{figure}

We describe the temporal dynamic convolution~(TDC) in this section, and its framework is shown in Figure~\ref{fig:framework-tdc}. Our main goal is to learn a set of coefficients to fuse the results of multiple convolutions with different kernel widths that provide plenty of temporal receptive field sizes. Therefore, TDC can adjust the temporal concept receptive field size dynamically based on the different input concept representations. 

Especially, given a type of concept representation $\boldsymbol{X}_{i}\in{\mathbb{R}^{L_i\times{N}}}$, 
and three 1-d convolution filters with different kernel widths~${K}_1\in{\mathbb{R}^{1}}, {K}_2\in{\mathbb{R}^{3}}, {K}_3\in{\mathbb{R}^{5}}$ that provide different temporal receptive fields, we separately perform three 1-d convolutions on $\boldsymbol{X}_i$ and obtain the feature representations  $\boldsymbol{X}_{i}^1\in{\mathbb{R}^{L_i\times{N}}}, \boldsymbol{X}_{i}^2\in{\mathbb{R}^{L_i\times{N}}}, \boldsymbol{X}_{i}^3\in{\mathbb{R}^{L_i\times{N}}}$ through, 
\begin{equation}
\boldsymbol{X}_i^j = Conv(\boldsymbol{X}_i, {K}_j), \ \  j\in{\left\{1,2,3
\right\}}
\end{equation}
where $Conv$ represents a 1-d convolution operation, $K_{j}$ is the parameters of the 1-d convolution filter ($j$ is the convolution kernel index).
In fact, we can utilize different numbers and sizes of convolution filters. We will give detail analysis in Section~\ref{subsection:ablation-study}.

Based on the results of these three convolutions, we employ a coefficient generation module to produce a set of coefficients and a results fusion module to fuse the convolution results. 

\subsubsection{Coefficient Generation Module}
The coefficient generation module is used to generate a set of coefficients based on the multiple convolution results, and its framework is shown in Figure~\ref{fig:framework-tdccoe}.

In order to generate the coefficients more flexible, we concatenate the results of all convolutions and apply a 1d convolution with kernel width 1 to reduce its channel number and obtain the feature representation $\boldsymbol{X}_{i}^{'}\in{\mathbb{R}^{L_{i}\times{N}}}$ through, 
\begin{equation}
\boldsymbol{X}_{i}^{'} = Conv(concat(\boldsymbol{X}_{i}^1, \boldsymbol{X}_{i}^2, \boldsymbol{X}_{i}^3), K_{1})
\end{equation}
where $Conv$ and $concat$ represent a 1-d convolution operation and concatenation operation, separately.

Based on $\boldsymbol{X}_{i}^{'}$, we will generate two sets of coefficients, which indicate the importance of different representations from two perspectives. They will be used to merge the multiple convolution results dynamically. It can assist us to capture appropriate and accurate temporal concept receptive field size for corresponding videos and highlight crucial concepts for efficient video recognition.

For one thing, we take into account the relations of all channels within the same convolution results and the relations of the same channel among all convolution results to generate a channel coefficient matrix $\boldsymbol{A}_{k}=[{a}_{k,1}, {a}_{k,2}, {a}_{k,3}]\in{\mathbb{R}^{{L_{i}}\times{3}}}$ through,
\begin{equation}
\boldsymbol{A}_{k} = softmax(tanh(\boldsymbol{X}_{i}^{'}\boldsymbol{W}_{k,1})\boldsymbol{W}_{k,2})
\end{equation}
where $\boldsymbol{W}_{k,1}\in{\mathbb{R}^{N\times{n}}}$, $\boldsymbol{W}_{k,2}\in{\mathbb{R}^{n\times{3}}}$ are learnable parameters.  $\boldsymbol{A}_{k}$ will be used to highlight the importance of different channels of each convolution results. For another, we consider all the convolution results and generate a time coefficient matrix $A_{t}\in{\mathbb{R}^{1\times{N}}}$ through, 
\begin{equation}
A_{t} = softmax(\boldsymbol{W}_{t,2}tanh(\boldsymbol{W}_{t,1}\boldsymbol{X}_{i}^{'}))
\end{equation}
where $\boldsymbol{W}_{t,1}\in{\mathbb{R}^{l\times{L_{i}}}}$, $\boldsymbol{W}_{t,2}\in{\mathbb{R}^{1\times{l}}}$ are learnable parameters. $A_{t}$ will be used to fuse the feature representation $\boldsymbol{X}_{i}^{''}$ through temporal dimension to capture the video level concept representation.

\subsubsection{Results Fusion Module}
The framework of results fusion models is shown in Figure~\ref{fig:framework-tdcres}.
Given the obtained convolution results $\left\{ \boldsymbol{X}_{i}^1, \boldsymbol{X}_{i}^2, \boldsymbol{X}_{i}^3 \right\}$ and generated coefficients matrix $\left\{ \boldsymbol{A}_{k}, A_{t} \right\}$, we first apply $\boldsymbol{A}_{k}$ on $\left\{ \boldsymbol{X}_{i}^1, \boldsymbol{X}_{i}^2, \boldsymbol{X}_{i}^3 \right\}$ and sum the results through channel dimension to obtain the feature representation $\boldsymbol{X}_{i}^{''}\in{\mathbb{R}^{L{i}\times{N}}}$ through,
\begin{equation}
\begin{split}
&\hat{x}_{i,t}^{j} = a_{k,j,t} \cdot {x}_{i,t}^j;  \ \ j\in{\left\{1,2,3\right\}}, t\in{\left\{1,2,..., L_{i}\right\}} \\
&\boldsymbol{X}_{i}^{''} = \sum_{j=1}^3 \boldsymbol{\hat{X}}_{i}^j\\
\end{split}
\end{equation}
where $a_{k,j}=[a_{k,j,1}; a_{k,j,2}; ... ; a_{k,j,L_{i}}]$, $\boldsymbol{X}_{i}^{j}=[{x}_{i,1}^{j}; {x}_{i,2}^{j}; ... ; {x}_{i,L_{i}}^{j}]$ and $\boldsymbol{\hat{X}}_{i}^{j}=[\hat{x}_{i,1}^{j}; \hat{x}_{i,2}^{j}; ... ; \hat{x}_{i,L_{i}}^{j}]\in{\mathbb{R}^{L_{i}\times{N}}}$. $\cdot$ is scalar-multiplication operation. After that, we capture the video level concept representation $X_{i}^{f}\in{\mathbb{R}^{1\times{L_{i}}}}$ by matrix multiplication through,
\begin{equation}
X_{i}^{f} = A_{t}\boldsymbol{X}_{i}^{''\top}
\end{equation}


\subsection{Temporal Dynamic Concept Modeling Network}
\label{tdcmn}

Based on the introduced temporal dynamic convolution, we propose the temporal dynamic concept modeling network~(TDCMN) for concept-based event recognition in this section. 

Our TDCMN consists of an intra-domain temporal dynamic concept modeling network~(InTDCM) and a cross-domain temporal dynamic concept modeling network~(CrTDCM), which utilizes TDC to capture the accurate and complete concept representation within and between different types of concepts. As described in Section~\ref{concept-based method}, we can obtain the initial concept representations  $\boldsymbol{X}_{1}\in{\mathbb{R}^{L_1\times{N}}}$ and $\boldsymbol{X}_{2}\in{\mathbb{R}^{L_2\times{N}}}$. We feed $\boldsymbol{X}_{1}$ and $\boldsymbol{X}_{2}$ into the InTDCM and CrTDCM to capture intra-domain dynamic concept representation  $X^{in}$ and cross-domain dynamic concept representation  $X^{cr}$. We will give the detail of InTDCM and CrTDCM in the following subsections. 

\subsubsection{Intra-domain Temporal Dynamic Concept Modeling Network}
\label{subsubsection:intdcmn}
In InTDCM, given $\left\{\boldsymbol{X}_{1}, \boldsymbol{X}_{2}\right\}$, we fed them into TDC separately and obtain the initial intra-domain dynamic concept representation $\left\{X_{1}^{in}, X_{2}^{in}\right\}$. Finally, we simply concatenate them and obtain the final intra-domain dynamic concept representation $X^{in}\in{\mathbb{R}^{1\times{L}}}$.

\subsubsection{Cross-domain Temporal Dynamic Concept Modeling Network}
\label{subsubsection:crtdcmn}
For CrTDCM, we exploit two methods to get the final cross-domain dynamic concept representation~(${X}_{si}^{cr}$ or ${X}_{co}^{cr}$) for event recognition. 

The simplest and most direct way is to concatenate the initial concept representations $\left\{\boldsymbol{X}_{1}, \boldsymbol{X}_{2}\right\}$ and feed it to the TDC. As described in Section~\ref{subsubsection:intdcmn}, we will obtain the final cross-domain dynamic concept representation ${X}_{si}^{cr}$. We term this method as CrTDCM$_{si}$.

\begin{figure}[t]
	\begin{center}
		\includegraphics[scale=0.45]{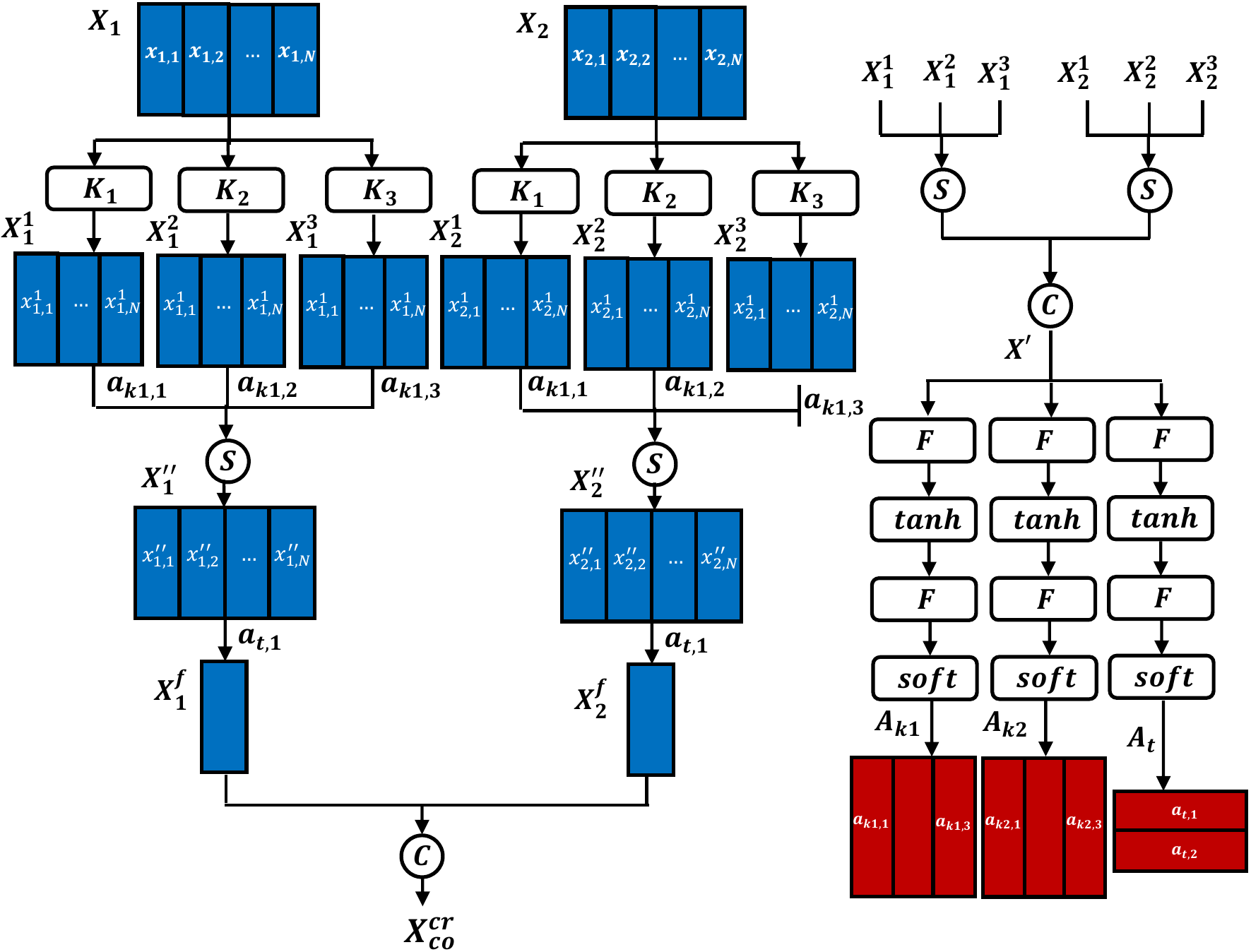}
	\end{center}
	\caption{Cross-domain temporal dynamic convolution.}
	\label{fig:framework-crtdcmnco}
\end{figure}

Furthermore, we construct a more efficient module to capture the cross-domain dynamic concept representation, and its framework is shown in Figure~\ref{fig:framework-crtdcmnco}. Specially, we feed $\boldsymbol{X}_{1}$ and $\boldsymbol{X}_{2}$ into three 1-d convolution layers with different kernel widths and obtain two types of concept representations $\left\{\boldsymbol{X}_{1}^1, \boldsymbol{X}_{1}^2, \boldsymbol{X}_{1}^3\right\}\in{\mathbb{R}^{L_1\times{N}}}$ and  $\left\{\boldsymbol{X}_{2}^1, \boldsymbol{X}_{2}^2, \boldsymbol{X}_{2}^3\right\}\in{\mathbb{R}^{L_2\times{N}}}$. Then we produce three sets of coefficients to fuse the convolution results dynamically and obtain the final cross-domain dynamic concept representation.

First, we fuse the concept representation of each type through element-wise summation and then concatenate them to obtain the feature representation $\boldsymbol{X}^{'}\in{\mathbb{R}^{L\times{N}}}$ for generating coefficients, 
\begin{equation}
\boldsymbol{X}^{'} = concat(sum(\boldsymbol{X}_{1}^1, \boldsymbol{X}_{1}^2, \boldsymbol{X}_{1}^3), sum(\boldsymbol{X}_{2}^1, \boldsymbol{X}_{2}^2, \boldsymbol{X}_{2}^3))
\end{equation}
where $sum$ and $concat$ represent element-wise summation and concatenation operations, separately.

And then we generate three sets of coefficients to fuse multiple convolution results based on $\boldsymbol{X}^{'}$. On the one hand, we generate two channel coefficient matrix $\boldsymbol{A}_{k1}\in{\mathbb{R}^{{L_{1}}\times{3}}}$ and $\boldsymbol{A}_{k2}\in{\mathbb{R}^{{L_{2}}\times{3}}}$ through, 
\begin{equation}
\begin{split}
&\boldsymbol{A}_{k1} = softmax(tanh(\boldsymbol{X}^{'}\boldsymbol{W}_{k1,1})\boldsymbol{W}_{k1,2})\\
&\boldsymbol{A}_{k2} = softmax(tanh(\boldsymbol{X}^{'}\boldsymbol{W}_{k2,1})\boldsymbol{W}_{k2,2})\\
\end{split}
\end{equation}
where $\boldsymbol{W}_{k1,1}\in{\mathbb{R}^{N\times{n}}}$, $\boldsymbol{W}_{k1,2}\in{\mathbb{R}^{n\times{3}}}$,  $\boldsymbol{W}_{k2,1}\in{\mathbb{R}^{N\times{n}}}$, $\boldsymbol{W}_{k2,2}\in{\mathbb{R}^{n\times{3}}}$ are learnable parameters. We take into account the relations within and between different types of concept representations when we calculate $\boldsymbol{A}_{k1}$ and $\boldsymbol{A}_{k2}$. They will be used to indicate the importance of different channels for each convolution results in different types of concepts, separately. On the other hand, we also generate a time coefficient matrix $\boldsymbol{A}_{t}\in{\mathbb{R}^{2\times{L_{i}}}}$ through, 
\begin{equation}
\boldsymbol{A}_{t} = softmax(\boldsymbol{W}_{t,2}tanh(\boldsymbol{W}_{t,1}\boldsymbol{X}^{'}))
\end{equation}
where $\boldsymbol{W}_{t,1}\in{\mathbb{R}^{l\times{L}}}$, $\boldsymbol{W}_{t,2}\in{\mathbb{R}^{2\times{l}}}$ are learnable parameters. $\boldsymbol{A}_{t}$ will be used to fuse the concept representation through temporal dimentation to get the video level concept representation.

Next, we apply $\boldsymbol{A}_{k1}$ and $\boldsymbol{A}_{k2}$ on $\left\{\boldsymbol{X}_{1}^1, \boldsymbol{X}_{1}^2, \boldsymbol{X}_{1}^3\right\}$ and  $\left\{\boldsymbol{X}_{2}^1, \boldsymbol{X}_{2}^2, \boldsymbol{X}_{2}^3\right\}$ to obtain the feature representation $\boldsymbol{X}_{1}^{''}\in{\mathbb{R}^{L_1\times{N}}}$ and $\boldsymbol{X}_{2}^{''}\in{\mathbb{R}^{L_2\times{N}}}$, separately,
\begin{equation}
\begin{split}
&\hat{x}_{1,t}^{j} = a_{k1,j,t} \cdot {x}_{1,t}^j;  \ \ j\in{\left\{1,2,3\right\}}, t\in{\left\{1,2,..., L_{i}\right\}} \\
&\hat{x}_{2,t}^{j} = a_{k2,j,t} \cdot {x}_{2,t}^j;  \ \ j\in{\left\{1,2,3\right\}}, t\in{\left\{1,2,..., L_{i}\right\}} \\
&\boldsymbol{X}_{1}^{''} = \sum_{j=1}^3 \boldsymbol{\hat{X}}_{1}^j; \ \ \boldsymbol{X}_{2}^{''} = \sum_{j=1}^3 \boldsymbol{\hat{X}}_{2}^j\\
\end{split}
\end{equation}
where $\boldsymbol{X}_{1}^{j}=[{x}_{1,1}^{j}; {x}_{1,2}^{j}; ... ; {x}_{1,L_{1}}^{j}]$, $\boldsymbol{X}_{2}^{j}=[{x}_{2,1}^{j}; {x}_{2,2}^{j}; ... ; {x}_{i,L_{2}}^{j}]$, $\boldsymbol{A}_{k1}=[{a}_{k1,1}, {a}_{k1,2}, {a}_{k1,3}]$, $\boldsymbol{A}_{k2}=[{a}_{k2,1}, {a}_{k2,2}, {a}_{k2,3}]$, 
$\boldsymbol{\hat{X}}_{1}^{j}=[\hat{x}_{1,1}^{j}; \hat{x}_{1,2}^{j}; ... ; \hat{x}_{1,L_{1}}^{j}]$, $\boldsymbol{\hat{X}}_{2}^{j}=[\hat{x}_{2,1}^{j}; \hat{x}_{2,2}^{j}; ... ; \hat{x}_{2,L_{2}}^{j}]$, $\cdot$ represents scalar-multiplication. Finally, we capture the final cross-domain dynamic concept representation $X_{co}^{cr}\in{\mathbb{R}^{1\times{L}}}$ through,
\begin{equation}
X_{co}^{cr} = concat(a_{t,1}\boldsymbol{X}_{1}^{''\top}, a_{t,2}\boldsymbol{X}_{2}^{''\top})
\end{equation}
where $\boldsymbol{A}_t=[{a}_{t,1}, {a}_{t,2}]$ and $concat$ represents concatenation operation. We term this method as CrTDCM$_{co}$.

\begin{figure}[t]
	\centering
	\subfigure[TDCMN$_{si}$]{
		\includegraphics[scale=0.4]{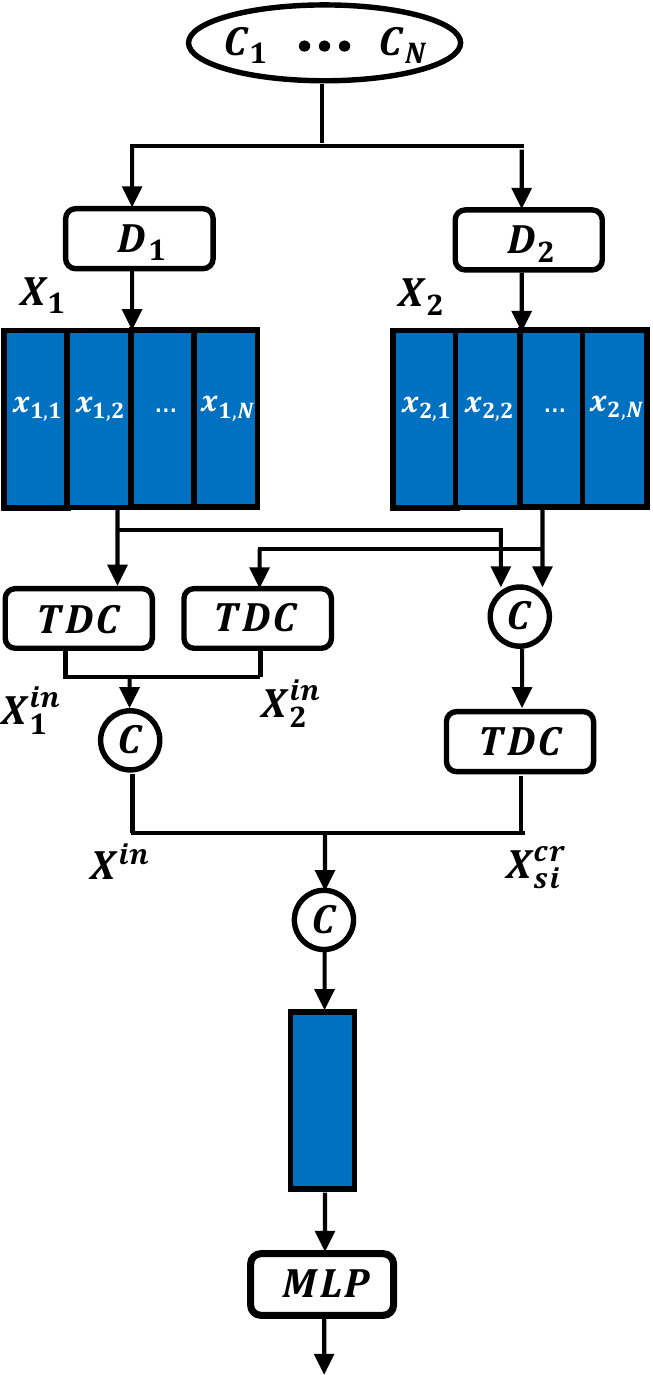}
		\label{fig:framework-tdcmnsi}
	}
	\quad
	\subfigure[TDCMN$_{co}$]{
		\includegraphics[scale=0.4]{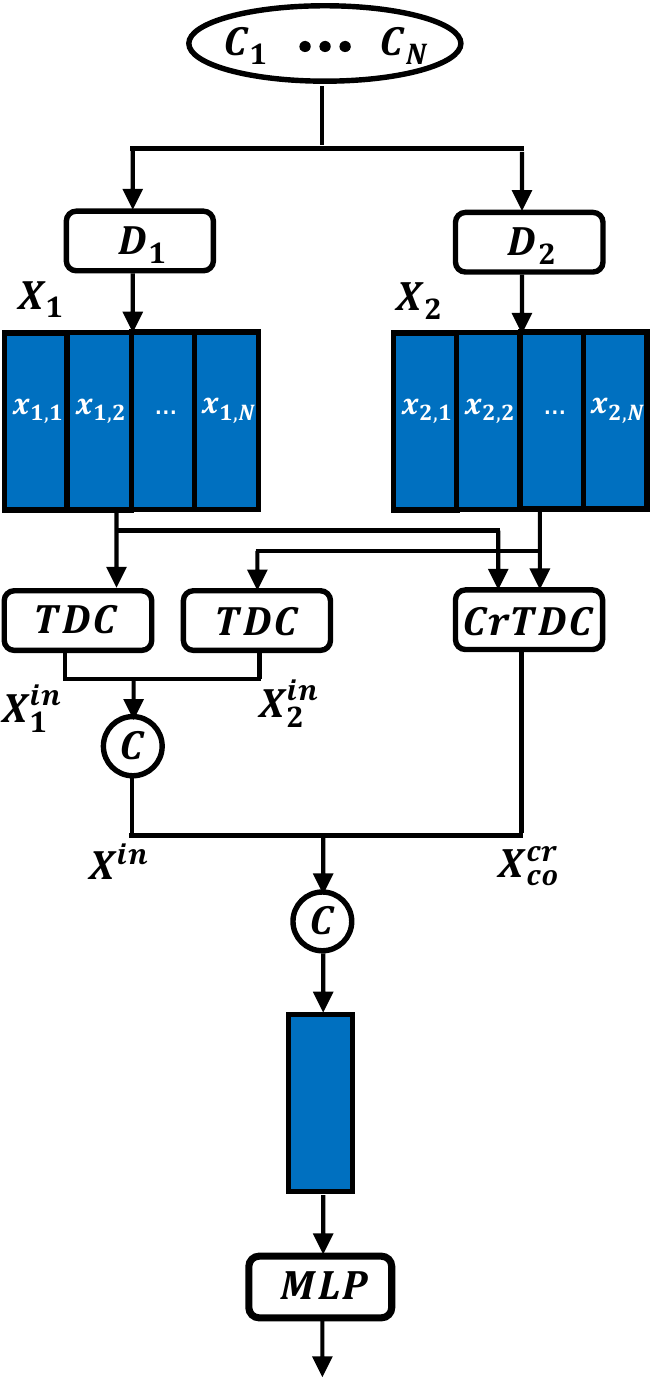}
		\label{fig:framework-tdcmnco}
	}
	\quad
	\centering
	\caption{Temporal dynamic concept modeling network.}
	\label{fig:framework}
\end{figure}

\subsubsection{Temporal Dynamic Concept Modeling Network}
As discussed in Section~\ref{subsubsection:crtdcmn}, we have constructed two cross-domain temporal dynamic concept modeling networks CrTDCM$_{si}$ and CrTDCM$_{co}$, therefore we have two temporal dynamic concept modeling networks, which we call them as TDCMN$_{si}$ and TDCMN$_{co}$, and the framework are shown in Figure~\ref{fig:framework-tdcmnsi} and Figure~\ref{fig:framework-tdcmnco}. For each model, we concatenate the obtained intra-domain and cross-domain dynamic concept representation and use an MLP network for final event recognition.

\section{Experiments}

\subsection{Experimental Setup}
\label{subsection:setting}

\textbf{Evaluation Datasets}: We use two large scale datasets for video event recognition: Fudan-Columbia Video Dataset (FCVID)\cite{jiang2018exploiting} and ActivityNet Dataset (ActivityNet)\cite{caba2015activitynet} in our experiments. The FCVID dataset contains 91,223 videos annotated with 239 event categories, including social events~({\it e.g.}, `tailgate party') and procedural events ({\it e.g.}, `making a cake'). Some videos can not be downloaded, and there are some corrupted videos that can not be used to extract video frames. We filter out these videos and end up with a training set of 45,416 videos and a testing set of 45,437 videos. The ActivityNet dataset is a large-scale untrimmed video dataset. We use the latest released version of the dataset (v1.3), which contains about 20K videos from 200 activity categories. Note that the annotations of the testing set have not been publicly available, we only use the training and validation set, and report results on the validation set.

\begin{table}[t]
	\renewcommand{\arraystretch}{1}
	\setlength{\tabcolsep}{11mm}
	\centering
	\caption{Ablation study about different types of concept detectors on FCVID.}
	\begin{tabular}{ c| c }
		\toprule[1.5pt]
		Model & mAP (\%)  \\
		\midrule
		Baseline$_{so}$ & 72.9 \\
		Baseline$_{soa}$ & 75.3 \\
		\midrule
		TDCMN$_{si,so}$ & 80.6 \\
		TDCMN$_{si,soa}$ & 81.9\\
		\midrule
		TDCMN$_{co,so}$ & 81.2 \\
		TDCMN$_{co,soa}$ & 82.2 \\
		\bottomrule
	\end{tabular}
	\label{table:detector-number}
	\vspace{-1ex}
\end{table}

\textbf{Evaluation Metric}: We compute average precision~(AP) for each event class and report the mean Average Precision (mAP) over all event categories. 

\textbf{Implementation Details}: We implement the proposed models based on the Pytorch framework. All proposed models are trained by SGD optimizer. The momentum and weight decay are set to 0.9 and 0.0005, respectively. We set the $N$ and $M$ as $16$ and $8$, respectively. For FCVID dataset, the batch size is set as 12. The initial learning rate is set to 0.5 and drops by 0.1 every 32 epochs. The training procedure stops after 40 epochs. For ActivityNet, the batch size is set as 12. The initial learning rate is set to 0.5 and drops by 0.1 every 40 epochs. The training procedure stops after 48 epochs. 

\subsection{The Selection and Use of Concept Detectors}
\label{subsection:choosen-detectors}

To guarantee the practicability of our method, we take into account several crucial elements when we pick concept detectors. First, in order to be practical, the concept detectors must be readily available. In other words, we can obtain them directly without training from scratch. Second, to guarantee the representation power of our model, different types of concept detectors must be employed. Third, the category of concepts included in each detector must be varied and universal, which assures the versatility of our model. Therefore, they can represent more event classes, and our model can be applied to more video event datasets. 

Based on the above considerations, we select the scene, object and action concept detectors, as these are the essential concepts for recognizing events. All these concept detectors are pre-trained on standard datasets and are readily available. We only use them to obtain initial concept representations. All layers of each detector except the final classification layer are fixed when we train our model. We will show how to use them below. 

\setlength{\parindent}{1em} \textbf{Scene Concept Detectors $D_{s}$.} ResNet-50 \cite{he2016deep} based $D_{s}$ pre-trained on Places365-Standard \cite{zhou2018places} dataset is utilized. $D_{s}$ contains 365 scene classes. The scene concept representation is extraced from the output of the last classification layer. Specially, an event video is evenly divided into $N$ clips and $M$ frames are randomly sampled from each clip. For each clip $C_n$, the concept representation $\boldsymbol{X}^{'}_{s, n}\in{\mathbb{R}^{365\times{M}}}$ is acquired by inputting $C_n$ into $D_s$. And then, the clip-level concept representation $X_{s, t}\in{\mathbb{R}^{365}}$ is obtained by imposing maximum pooling on $\boldsymbol{X}^{'}_{s, n}$ over all frames. Lastly, the video-level initial scene concept representation $\boldsymbol{X}_{s}\in{\mathbb{R}^{365\times{N}}}$ is obtained by concatenating the representations of all the video clips.

\begin{table}[t]
	\renewcommand{\arraystretch}{1}
	\setlength{\tabcolsep}{1.5mm}
	\centering
	\caption{Ablation study about modeling intra-domain and cross-domain concept representation on FCVID.}
	\begin{tabular}{ c| c| c | c | c}
		\toprule[1.5pt]
		Model & InTDCM & CrTDCM$_{si}$ & CrTDCM$_{co}$ & mAP (\%) \\
		\midrule
		Baseline$_{so}$ & & & & 72.9 \\
		+InTDCM & \checkmark & & & 77.9 \\
		\midrule
		+CrTDCM$_{si}$ & & \checkmark &  & 78.4 \\
		TDCMN$_{si,so}$ & \checkmark & \checkmark & & \textbf{ 80.6 }\\
		\midrule
		+CrTDCM$_{co}$ & &  & \checkmark & 77.6 \\
		TDCMN$_{co,so}$ & \checkmark & & \checkmark & \textbf{ 81.2 }\\
		\bottomrule
	\end{tabular}
	\label{table:ablation-study}
	\vspace{-1ex}
\end{table}

\textbf{Object Concept Detectors $D_{o}$.} We use ResNet-50 based object concept detector $D_{o}$ pre-trained on ImageNet \cite{russakovsky2015imagenet} as $D_{o}$. Similarly, we obtain the video-level initial object concept representation $\boldsymbol{X}_{o}\in{\mathbb{R}^{1000\times{N}}}$, where $1000$ is the number of object concepts in $D_{o}$.

\textbf{Action Concept Detectors $D_{a}$.} We use the I3D-based~\cite{carreira2017quo}  $D_{a}$ pre-trained on kinetics \cite{kay2017kinetics} dataset. There are 400 action classes in $D_{a}$. We also employ the output of the last classification layer as the action concept representation. We input each video clip $C_n$ to $D_{a}$ and get the clip-level action concept representation $X_{a,n}\in{\mathbb{R}^{400}}$. Finally, we concatenate the representation of all clips and obtain the video-level initial action concept representation $\boldsymbol{X}_{a}\in{\mathbb{R}^{{400}\times{N}}}$.

\subsection{Ablation Studies}
\label{subsection:ablation-study}

In this section, we construct abundant experiments to testify the efficiency of each choice of our model. 

\subsubsection{\textbf{The number of concept detectors}}
\label{subsection:detector-number}
The video comprises a variety of concepts, which work together to describe the event of this video. In this subsection, we conduct six experiments on FCVID to show the efficiency of different types of concepts. The details setting are shown below, and the results are shown in Table~\ref{table:detector-number}. 

We only use scene and object concept detectors in Baseline$_{so}$. We concatenate the results of these two detectors and use a fully connected layer to perform event recognition. Baseline$_{soa}$ is similar to Baseline$_{so}$ except that the action concept detectors are also used. TDCMN$_{si,so}$ is TDCMN$_{si}$ with scene and object concept detectors. TDCMN$_{si,soa}$ is TDCMN$_{si}$ with scene, object and action concept detectors. TDCMN$_{co,so}$ is TDCMN$_{co}$ with use scene and object concept detectors. TDCMN$_{co,soa}$ is TDCMN$_{co}$ with scene, object and action concept detectors.

By comparing the results of Baseline$_{so}$ and Baseline$_{soa}$~(or the results of TDCMN$_{si,so}$ and TDCMN$_{si,soa}$ or the results of TDCMN$_{co,so}$ and TDCMN$_{co,soa}$), we can find that the event recognition performance is higher when more concept detectors are used. This is because more concept detectors can give us richer concept representations, which can describe the event more comprehensiveness and accuracy. By comparing the results of our model with baseline methods, we can find that the efficiency of our model is testified no matter how many concept detectors are used. Therefore, we will only use scene and object concept detectors in the following experiments of this section.

\begin{table}[t]
	\renewcommand{\arraystretch}{1}
	\setlength{\tabcolsep}{11mm}
	\centering
	\caption{Ablation study about temporal dynamic convolution on FCVID.}
	\begin{tabular}{ c| c }
		\toprule[1.5pt]
		Model & mAP (\%) \\
		\midrule
		w/o TDC &  76.0 \\
		with TDC & \textbf{77.9} \\
		\bottomrule
	\end{tabular}
	\label{table:tdc}
	\vspace{-1ex}
\end{table}

\begin{table}[t]
	\renewcommand{\arraystretch}{1}
	\setlength{\tabcolsep}{4mm}
	\centering
	\caption{Ablation study about convolution filters with different kernel widths on FCVID. $K_1$, $K_2$, $K_3$ and $K_4$ represent 1-d convolution filters with kernel width $1$, $3$, $5$ and $7$, separately.}
	\begin{tabular}{ c| c| c | c| c | c}
		\toprule[1.5pt]
		Exp. & $K_1$ & $K_2$ & $K_3$ & $K_4$ & mAP (\%) \\
		\midrule
		1 & \checkmark & \checkmark & & & 76.8 \\
		2 & \checkmark & & \checkmark & & 76.7 \\
		3 & & \checkmark & \checkmark & &  77.4 \\
		4 & \checkmark & \checkmark & \checkmark & & \textbf{77.9} \\
		5 & \checkmark & \checkmark & \checkmark & \checkmark & 77.8 \\
		\bottomrule
	\end{tabular}
	\label{table:kernel-number}
	\vspace{-1ex}
\end{table}

\subsubsection{\textbf{The necessary of modeling intra-domain and cross-domain concept representations}}
\label{subsection:in-cross-necessary}
We propose two temporal dynamic concept modeling networks TDCMN$_{si}$ and TDCMN$_{co}$. Each network models both intra-domain and cross-domain concept representations. We construct experiments to demonstrate the necessity of modeling intra-domain and cross-domain concept representations, and the results are shown in Table~\ref{table:ablation-study}. We can see that the InTDCM and the CrTDCM$_{si}$~(or CrTDCM$_{co}$) both can improve the event recognition performance to some extent compared to Baseline$_{so}$, which indicates the efficiency of each module. Finally, the event recognition performance has further promoted by modeling the intra-domain and cross-domain concept representations at the same time in TDCMN$_{si,so}$~(or TDCMN$_{co,so}$). This phenomenon verifies the necessary and complementarity of modeling intra-domain and cross-domain concept representations. Hence, we will only conduct experiments on InTDCM in the following subsections.

\subsubsection{\textbf{Temporal dynamic convolution}}
\label{subsection:tdc}
The core of this paper is the proposed temporal dynamic convolution, we prepare experiments to verify its effectiveness, and the results are shown in Table~\ref{table:tdc}. According to the previous analysis, we construct experiments on InTDCM to see the performance change when we remove the temporal dynamic convolution from it. From Table~\ref{table:tdc}, we can see that the event recognition performance dropped by 1.9\% mAP, which proves the effectiveness of the temporal dynamic convolution. The proposed TDC can generate two sets of coefficients, which indicates the importance of different channels within and between each convolution results and the importance of different video clips. Therefore, we can obtain crucial concepts through TDC.

\subsubsection{\textbf{The number of convolution filters with different kernel widths}}
\label{subsubsection:kernel-number}
Up to now, we verify the necessity of modeling intra-domain and cross-domain concept representations and the effectiveness of temporal dynamic convolution. Therefore, we construct experiments on InTDCM to see the event recognition performance change when a different number of convolution filters with different kernel widths are used. The results are shown in Table~\ref{table:kernel-number}. We can find that event recognition performance is increased when we use more convolution filters. It is mainly because that convolution with different kernel widths can provide different sizes of temporal concept receptive field. We can obtain various temporal receptive field sizes for different concepts. Hence, we can find the appropriate temporal concept receptive field size and capture useful concepts for event recognition. Besides, by comparing the results of Exp.4 and Exp.5 in Tabel~\ref{table:kernel-number}, we can find a slight drop in performance when using too many convolution kernels. Therefore, we only use three convolution filters with kernel widths $1,3,5$ in all experiments.

\begin{table}[t]
	\renewcommand{\arraystretch}{1}
	\setlength{\tabcolsep}{4mm}
	\centering
	\caption{The comparison between our method and other Concept-based methods on FCVID and ActivityNet dataset.}
	\begin{tabular}{c|c|c}
		\toprule[1.5pt]
		Method & FCVID & ActivityNet \\
		\midrule
		Early Fusion-NN  \cite{wu2016harnessing} & 75.2 & 55.9 \\
		Fusion-SVM \cite{wu2016harnessing} & 75.5 & 55.8 \\
		SVM-MKL \cite{kloft2011lp} & 74.9 & 56.3 \\
		OSF \cite{wu2016harnessing} & 76.5 & 56.8 \\ 
		\midrule
		TDCMN$_{si,soa}$ & 81.9 &  84.3 \\
		TDCMN$_{co,soa}$ & \textbf{82.2} & \textbf{84.6} \\
		\bottomrule
	\end{tabular}
	\label{table:compare-results-concept}
	\vspace{-1ex}
\end{table}

\begin{table}[t]
	\renewcommand{\arraystretch}{1}
	\setlength{\tabcolsep}{1.2mm}
	\centering
	\caption{The comparison between our method and other Appearance-based methods on FCVID and ActivityNet.}
	\begin{tabular}{c | c |c |c}
		\toprule[1.5pt]
		Method & FCVID & Method & ActivityNet \\
		\midrule
		DMF \cite{smith2003multimedia} & 72.5 & AdaFrame \cite{wu2019adaframe} & 71.57 \\
		DASD \cite{jiang2012fast} & 72.8 & LiteEval \cite{wu2019liteeval} & 72.7 \\
		M-DBM \cite{srivastava2012multimodal} & 74.4 & TSN \cite{wang2016temporal} & 76.6 \\
		rDNN \cite{jiang2018exploiting} & 76.0 & KeylessAttention \cite{long2018multimodal} & 78.5 \\
		GSFMN-all \cite{zhao2019visual} & 76.9 & P3D \cite{qiu2017learning} & 78.9 \\ 
		Pivot CorrNN \cite{kang2018pivot} & 77.6 & MLSME \cite{zhang2019exploiting} & 83.0 \\
		LiteEval \cite{wu2019liteeval} & 80.0 & MARL-based \cite{wu2019multi} & 83.8 \\
		AdaFrame \cite{wu2019adaframe} & 80.2 & IMGAUD2VID \cite{gao2019listen} & 84.2 \\ 
		\midrule
		TDCMN$_{si,soa}$ & 81.9 & TDCMN$_{si,soa}$ & 84.3 \\
		TDCMN$_{co,soa}$ & \textbf{82.2} & TDCMN$_{co,soa}$ & \textbf{84.6} \\
		\bottomrule
	\end{tabular}
	\label{table:compare-results-appearance}
	\vspace{-1ex}
\end{table}

\subsection{Comparison to start of the art}
\label{subsection:comparison}

We compare our model with both Appearance-based methods and Concept-based methods on FCVID and ActivityNet datasets. The results are shown in Table~\ref{table:compare-results-concept} and Table~\ref{table:compare-results-appearance}.

\textbf{Concept-based Models.} 
The comparison results between our method and other Concept-based methods on FCVID and ActivityNet datasets are shown in Table~\ref{table:compare-results-concept}. As can be seen from the table, our models can improve the performance of Concept-based event recognition models by a large margin, especially on ActivityNet. Particularly, our model is at least 5.4\% mAP higher than OSF on FCVID. Though OSF uses vgg features and resnet are used in our model, OSF utilizes the generic vision feature that directly extracted from videos and more than 20000 concepts. Besides, OSF uses average pooling over the concept representations of all video clips and ignores the temporal characteristic of concept representations, which causes performance reduction. 

\begin{figure*}[t]
	\centering
	\subfigure[`Beach', `scene', `K2']{
		\includegraphics[scale=0.25]{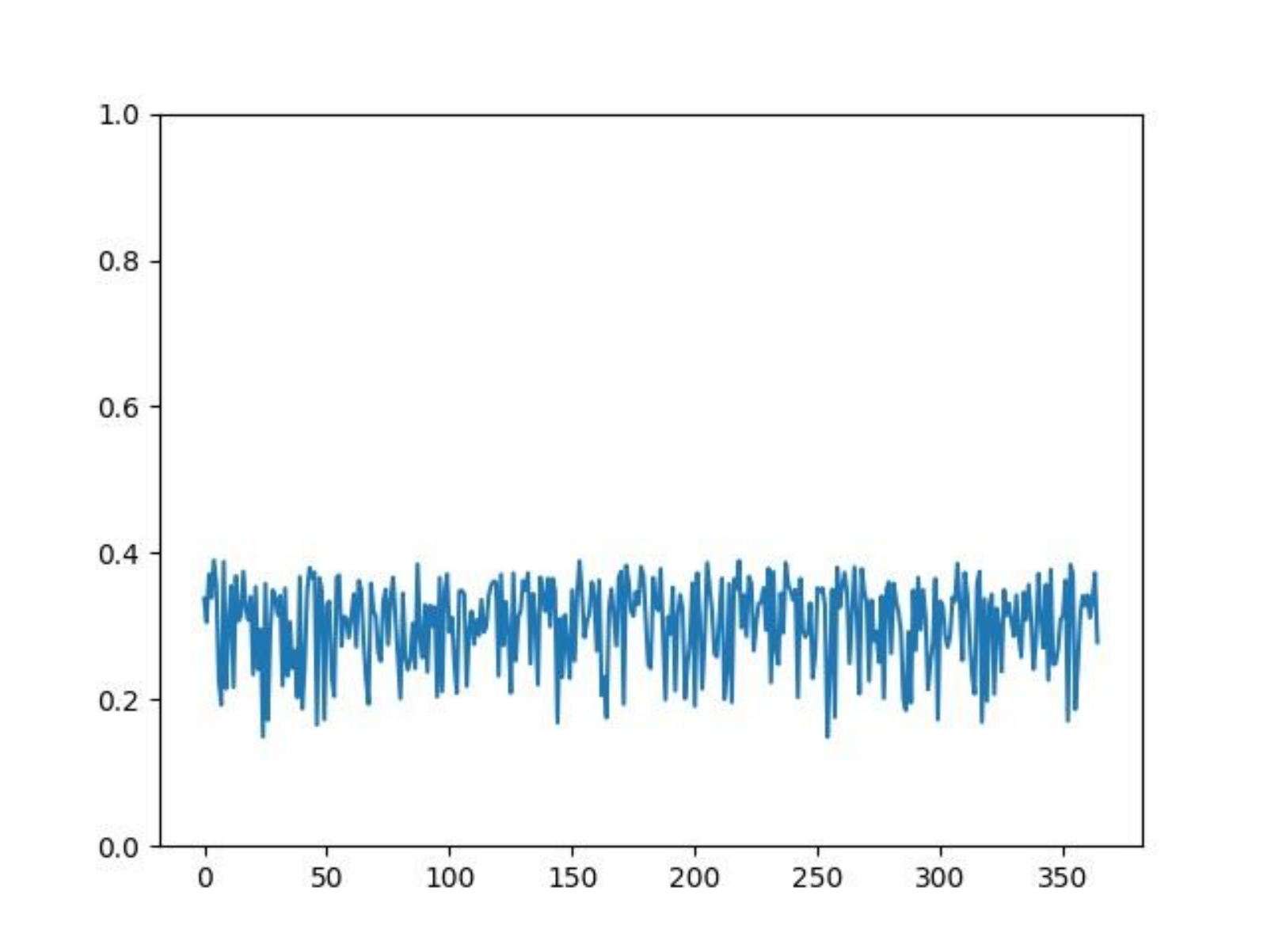}
		\label{fig:tdc-visualize.1}
	}
	\quad
	\hspace{-2ex}
	\vspace{-0.5ex}
	\subfigure[`Beach', `scene', `K3']{
		\includegraphics[scale=0.25]{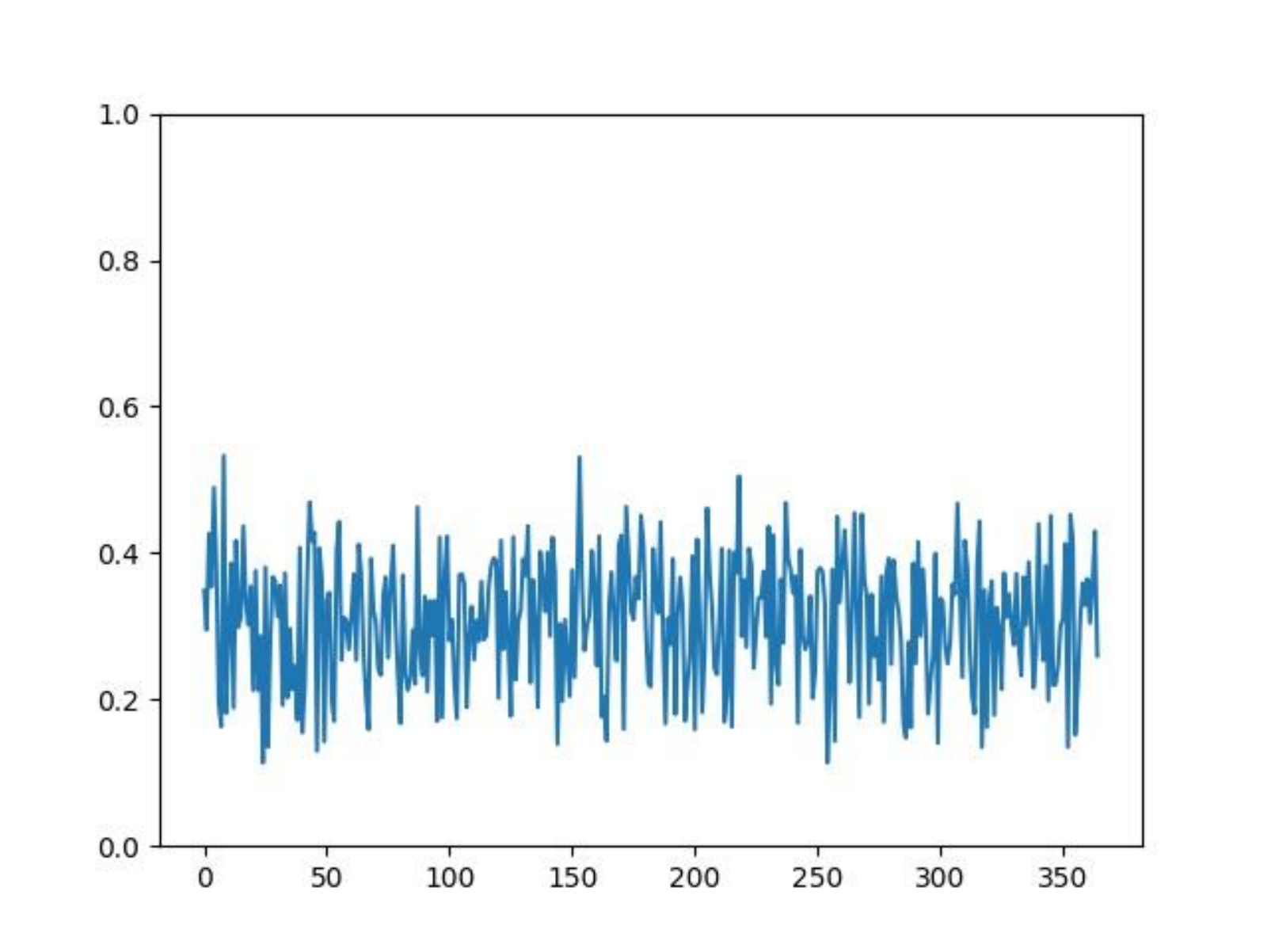}
		\label{fig:tdc-visualize.2}
	}
	\quad
	\hspace{-2ex}
	\vspace{-0.5ex}
	\subfigure[`Birthday party', `object', `K2']{
		\includegraphics[scale=0.25]{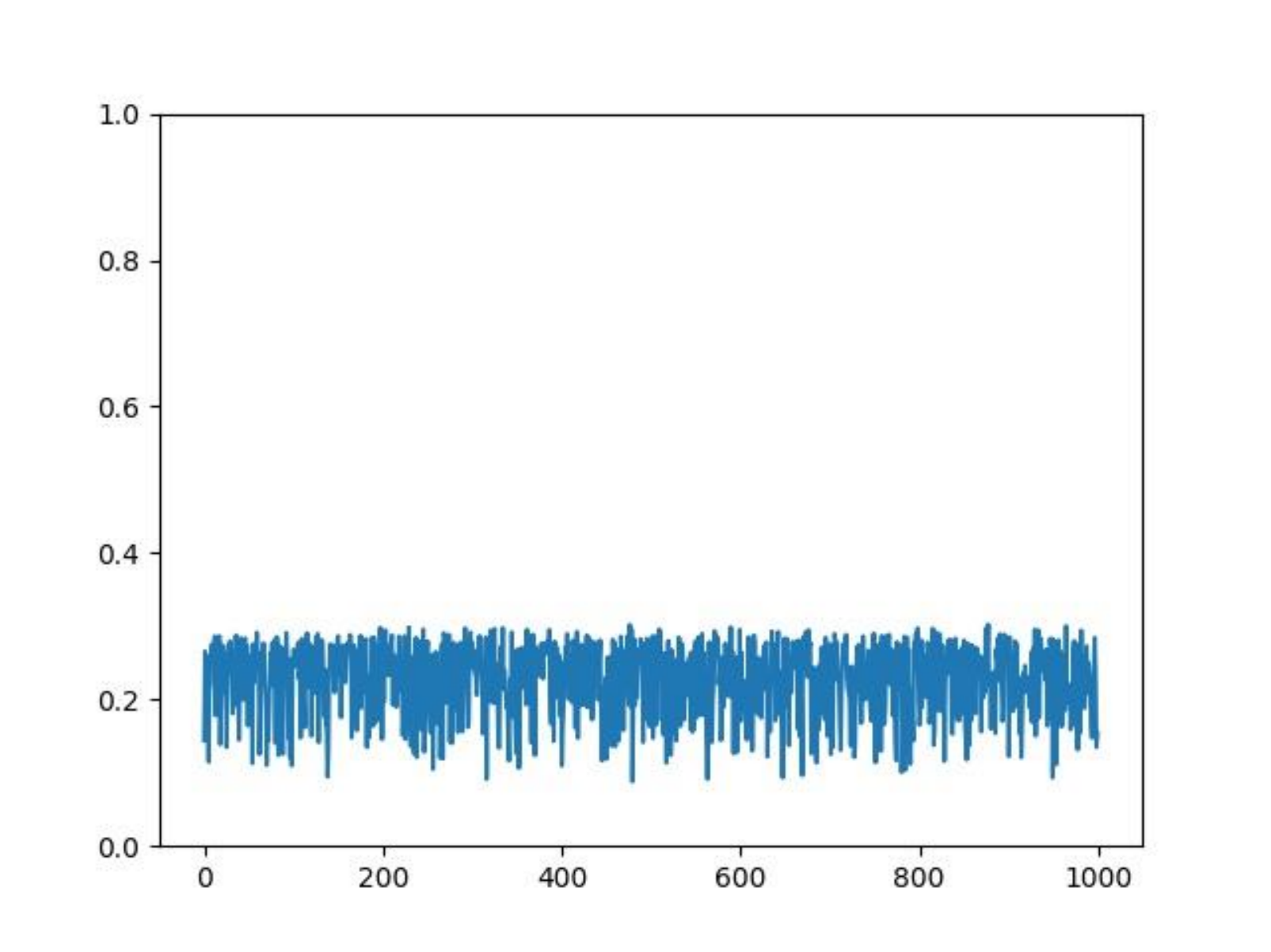}
		\label{fig:tdc-visualize.3}
	}
	\quad
	\hspace{-2ex}
	\vspace{-0.5ex}
	\subfigure[`Birthday party', `object', `K3']{
		\includegraphics[scale=0.25]{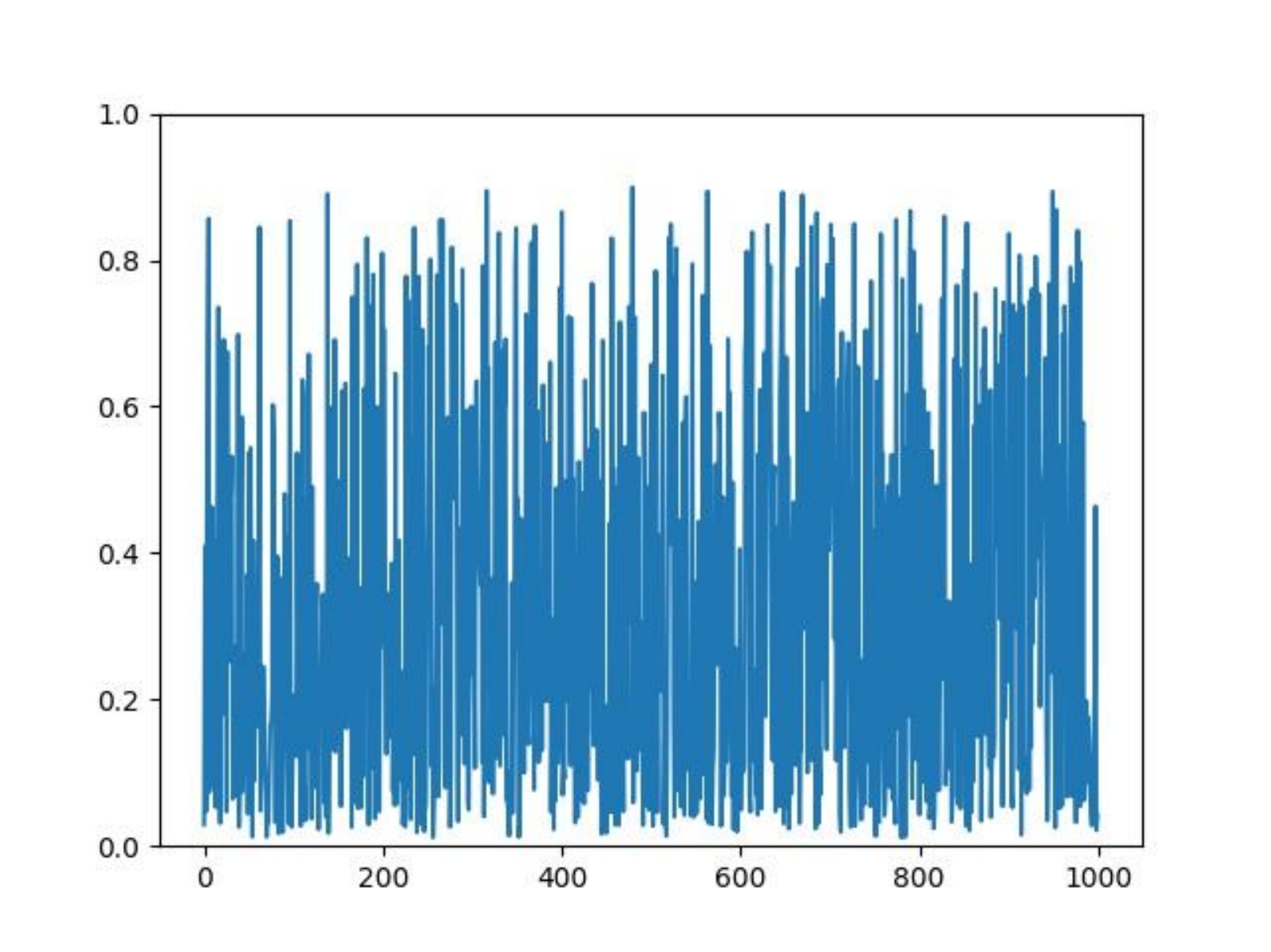}
		\label{fig:tdc-visualize.4}
	}
	\quad
	\hspace{-2ex}
	\vspace{-0.5ex}
	\subfigure[`All', `scene', `K2-K1']{
		\includegraphics[scale=0.25]{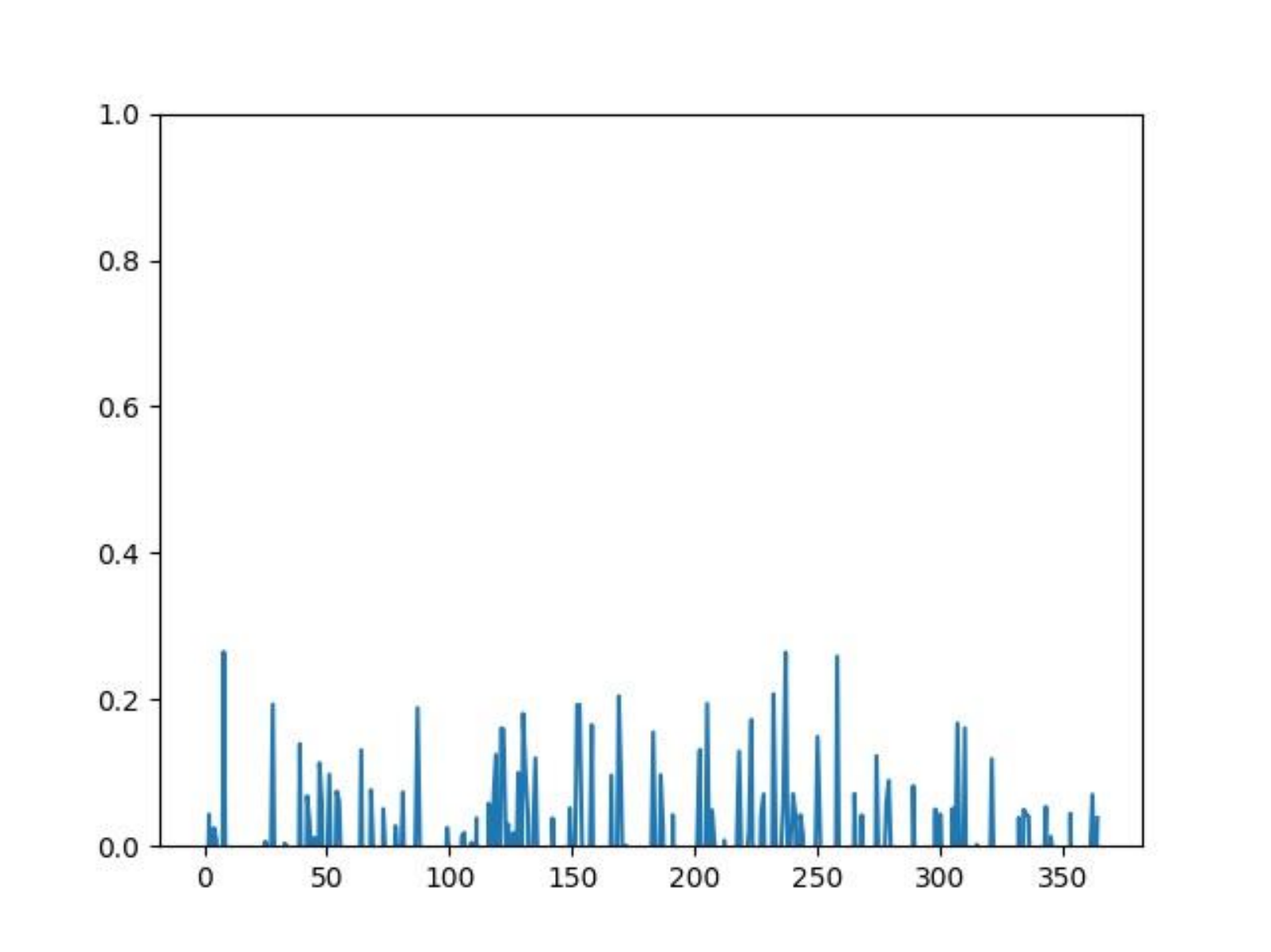}
		\label{fig:tdc-visualize.5}
	}
	\quad
	\hspace{-2ex}
	\subfigure[`All', `object', `K2-K1']{
		\includegraphics[scale=0.25]{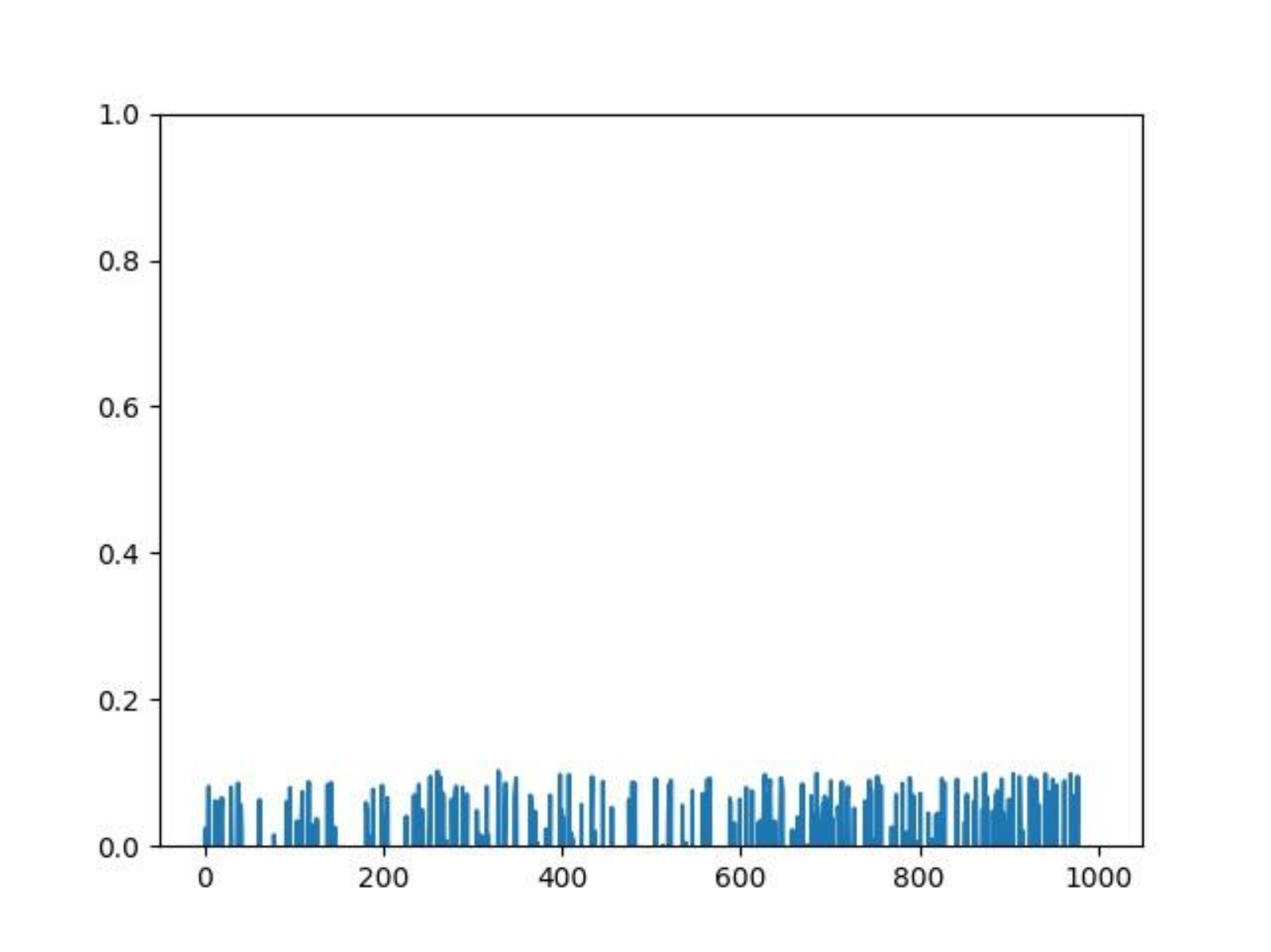}
		\label{fig:tdc-visualize.6}
	}
	\quad
	\hspace{-2ex}
	\subfigure[`All', `scene', `K3-K2']{
		\includegraphics[scale=0.25]{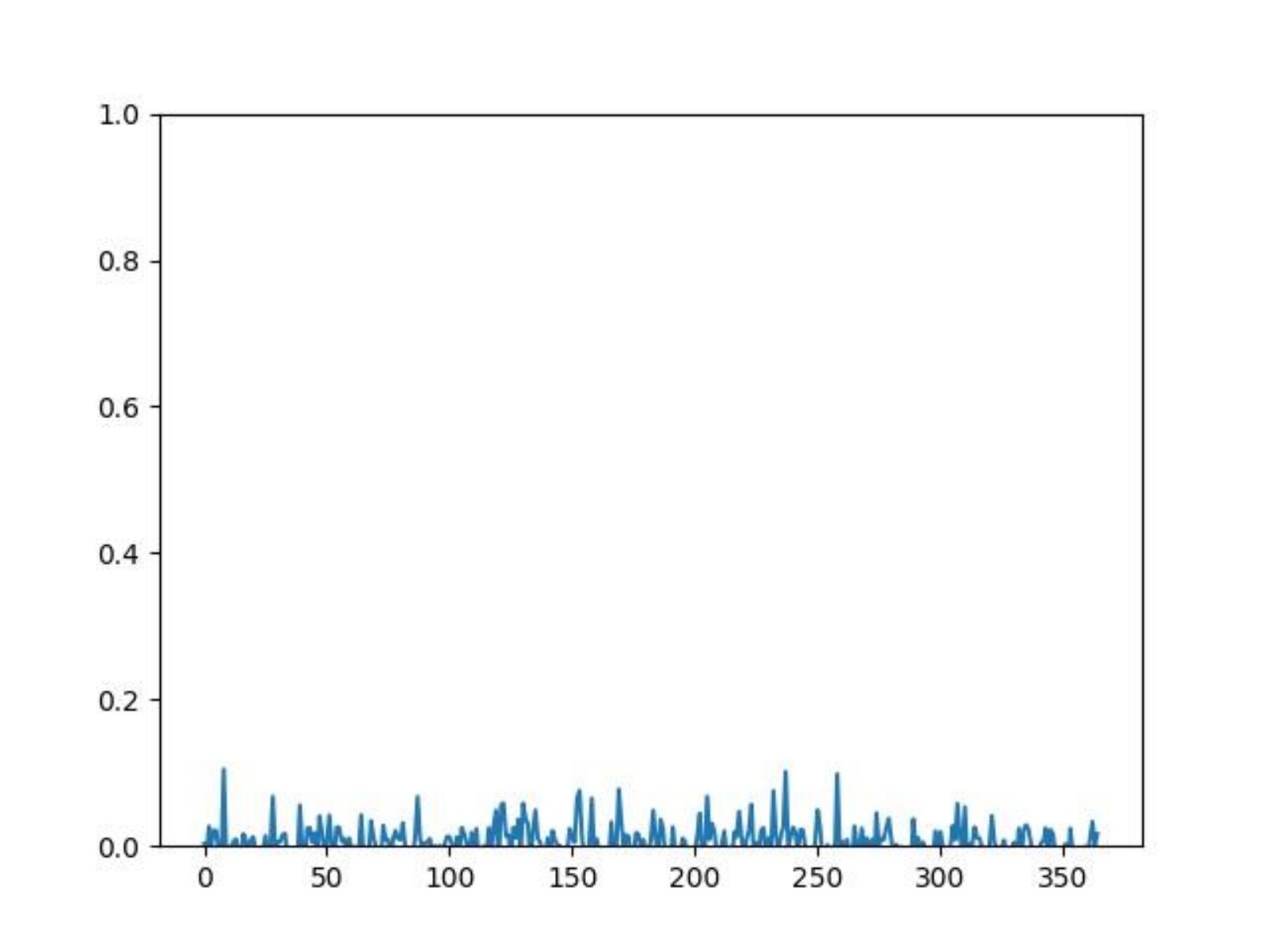}
		\label{fig:tdc-visualize.7}
	}
	\quad
	\hspace{-2ex}
	\subfigure[`All', `object', `K3-K2']{
		\includegraphics[scale=0.25]{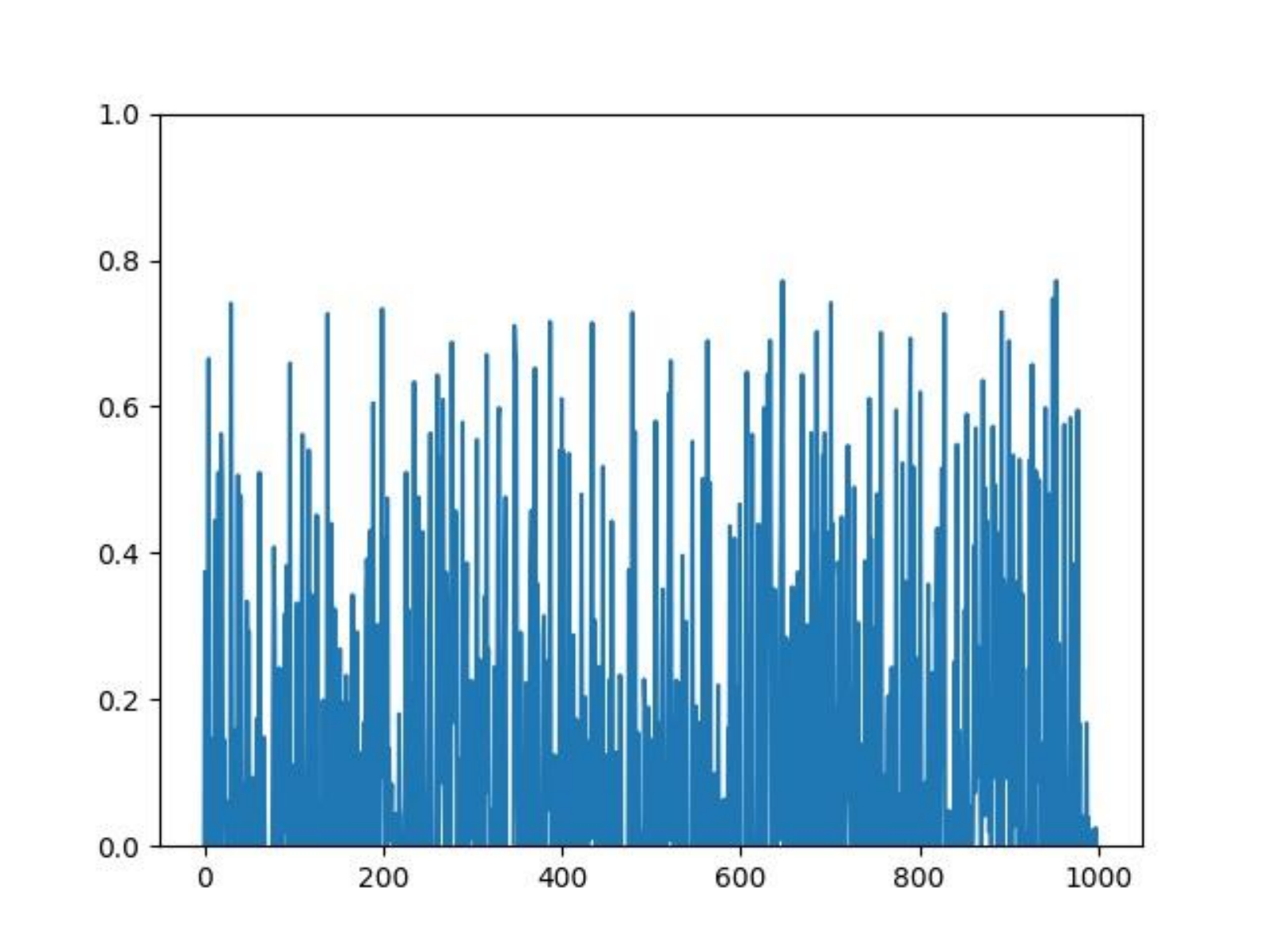}
		\label{fig:tdc-visualize.8}
	}
	\quad
	\hspace{-2ex}
	\centering
	\vspace{-2ex}
	\caption{Visualized the distribution of coefficients generated by temporal dynamic convolution in InTDCM module. `Beach', `scene', `K2' means this line graph shows the channel coefficient distribution generated for convolution kernel `K2' of `scene' concept on `Beach' category. `All', `scene', `K2-K1' means this line graph shows the difference of the channel coefficients between convolution kernel `K2' and `K1' of `scene' concept on all event categories. The other subgraphs have similar meanings.}
	\label{fig:tdc-visualize}
	\vspace{-2ex}
\end{figure*}

\textbf{Appearance-based Models.}
The comparison results between our method and some Appearance-based methods on FCVID and ActivityNet datasets are shown in Table~\ref{table:compare-results-appearance}. We can find that our model achieves higher event recognition performance compared with some Appearance-based methods on both datasets. Besides, our model is also better than Pivot CorrNN~\cite{kang2018pivot}, which uses seven types of pre-extracted features to perform event recognition.

\subsection{Visualization}
\label{subsection:visualization}
To understand the temporal dynamic convolution more clearly, we visualize the distribution of the coefficients generated by TDC in InTDCM module, and the results are shown in Figure~\ref{fig:tdc-visualize}. On the one hand, we seek to see the behavior of TDC from the perspective of event categories. Especially, we randomly sample two event categories `Beach' and `Birthday party'. Then, we calculate the average of the channel coefficients generated for each convolution filters across all validation videos in each category. From Figure~\ref{fig:tdc-visualize.1} and \ref{fig:tdc-visualize.2}, we can find that the channel coefficient distribution of convolutions with different kernel widths is similar for scene concept of `Beach' event category. Instead, the channel coefficient distribution is quite different for object concept of `Birthday party'~(see Figure~\ref{fig:tdc-visualize.3} and \ref{fig:tdc-visualize.4}). This phenomenon is consistent with our motivation. For `Beach', the temporal existence patterns of scene concepts at different time scales are similar. Therefore, the coefficient distribution for different convolution results are similar. In contrast, the temporal existence patterns of object concepts at different time scales are quite different for `Birthday party'. Hence, the needed temporal receptive field size is different. Accordingly, the learned coefficients of each convolution result are different. 

On the other hand, we also exploit the behavior of TDC from the perspective of concept types. Notably, we calculate the difference of the channel coefficients generated for different convolution filters of each type of concept across all validation videos in all event categories. From Figure~\ref{fig:tdc-visualize.5} and \ref{fig:tdc-visualize.6}, we can find that the difference between K2 and K1 are similar for scene and object concepts. In contrast, the difference between K3 and K2 are different for scene and object concepts by comparing Figure~\ref{fig:tdc-visualize.7} and \ref{fig:tdc-visualize.8}. This is mainly because different types of concepts have its unique temporal existence pattern. Our model can generate corresponding temporal concept receptive field sizes adaptively based on different types of concepts.
\section{Conclusion}
In this paper, we explore the temporal receptive field of concept-based event recognition methods for efficiently untrimmed video event analysis. First, we introduce a temporal dynamic convolution~(TDC) to give stronger flexibility to concept-based event recognition networks, which can adjust its receptive field size adaptively based on different inputs. Based on TDC, we propose the temporal dynamic concept modeling network~(TDCMN) to learn an accurate and complete concept representation for efficient untrimmed video analysis. TDCMN employs TDC to analyze the temporal characteristic of concepts within the same type and between different types. To demonstrate the effectiveness of our model, we apply TDCMN on two challenging video datasets FCVID and ActivityNet. TDCMN can improve event recognition performance by a large margin compared with other concept-based event recognition methods.

\begin{acks}
This work was supported in part by the National Key R\&D Program of China under Grant 2018AAA0102003 and 2018YFE0118400, in part by National Natural Science Foundation of China: 61672497, 61620106009, 61836002, 61931008, 61976069, 61650202 and U1636214, and in part by Key Research Program of Frontier Sciences, CAS: QYZDJ-SSW-SYS013. We acknowledge Kingsoft Cloud for the helpful discussion and free GPU cloud computing resource support. 
\end{acks}

\vfil\eject
\bibliographystyle{ACM-Reference-Format}
\bibliography{sample-base}


\begin{thebibliography}{47}


\ifx \showCODEN    \undefined \def \showCODEN     #1{\unskip}     \fi
\ifx \showDOI      \undefined \def \showDOI       #1{#1}\fi
\ifx \showISBNx    \undefined \def \showISBNx     #1{\unskip}     \fi
\ifx \showISBNxiii \undefined \def \showISBNxiii  #1{\unskip}     \fi
\ifx \showISSN     \undefined \def \showISSN      #1{\unskip}     \fi
\ifx \showLCCN     \undefined \def \showLCCN      #1{\unskip}     \fi
\ifx \shownote     \undefined \def \shownote      #1{#1}          \fi
\ifx \showarticletitle \undefined \def \showarticletitle #1{#1}   \fi
\ifx \showURL      \undefined \def \showURL       {\relax}        \fi
\providecommand\bibfield[2]{#2}
\providecommand\bibinfo[2]{#2}
\providecommand\natexlab[1]{#1}
\providecommand\showeprint[2][]{arXiv:#2}

\bibitem[\protect\citeauthoryear{Bhattacharya, Kalayeh, Sukthankar, and
  Shah}{Bhattacharya et~al\mbox{.}}{2014}]%
        {bhattacharya2014recognition}
\bibfield{author}{\bibinfo{person}{Subhabrata Bhattacharya},
  \bibinfo{person}{Mahdi~M Kalayeh}, \bibinfo{person}{Rahul Sukthankar}, {and}
  \bibinfo{person}{Mubarak Shah}.} \bibinfo{year}{2014}\natexlab{}.
\newblock \showarticletitle{Recognition of complex events: Exploiting temporal
  dynamics between underlying concepts}. In
  \bibinfo{booktitle}{\emph{Proceedings of the IEEE conference on computer
  vision and pattern recognition}}. \bibinfo{pages}{2235--2242}.
\newblock


\bibitem[\protect\citeauthoryear{Burkov and Lempitsky}{Burkov and
  Lempitsky}{2018}]%
        {burkov2018deep}
\bibfield{author}{\bibinfo{person}{Egor Burkov} {and} \bibinfo{person}{Victor
  Lempitsky}.} \bibinfo{year}{2018}\natexlab{}.
\newblock \showarticletitle{Deep Neural Networks with Box Convolutions}.
\newblock  (\bibinfo{year}{2018}), \bibinfo{pages}{6211--6221}.
\newblock


\bibitem[\protect\citeauthoryear{Caba~Heilbron, Escorcia, Ghanem, and
  Carlos~Niebles}{Caba~Heilbron et~al\mbox{.}}{2015}]%
        {caba2015activitynet}
\bibfield{author}{\bibinfo{person}{Fabian Caba~Heilbron},
  \bibinfo{person}{Victor Escorcia}, \bibinfo{person}{Bernard Ghanem}, {and}
  \bibinfo{person}{Juan Carlos~Niebles}.} \bibinfo{year}{2015}\natexlab{}.
\newblock \showarticletitle{Activitynet: A large-scale video benchmark for
  human activity understanding}. In \bibinfo{booktitle}{\emph{Proceedings of
  the IEEE Conference on Computer Vision and Pattern Recognition}}.
  \bibinfo{pages}{961--970}.
\newblock


\bibitem[\protect\citeauthoryear{Carreira and Zisserman}{Carreira and
  Zisserman}{2017}]%
        {carreira2017quo}
\bibfield{author}{\bibinfo{person}{Joao Carreira} {and} \bibinfo{person}{Andrew
  Zisserman}.} \bibinfo{year}{2017}\natexlab{}.
\newblock \showarticletitle{Quo vadis, action recognition? a new model and the
  kinetics dataset}. In \bibinfo{booktitle}{\emph{proceedings of the IEEE
  Conference on Computer Vision and Pattern Recognition}}.
  \bibinfo{pages}{6299--6308}.
\newblock


\bibitem[\protect\citeauthoryear{Chang, Yu, Yang, and Hauptmann}{Chang
  et~al\mbox{.}}{2015}]%
        {chang2015searching}
\bibfield{author}{\bibinfo{person}{Xiaojun Chang}, \bibinfo{person}{Yao-Liang
  Yu}, \bibinfo{person}{Yi Yang}, {and} \bibinfo{person}{Alexander~G
  Hauptmann}.} \bibinfo{year}{2015}\natexlab{}.
\newblock \showarticletitle{Searching persuasively: Joint event detection and
  evidence recounting with limited supervision}. In
  \bibinfo{booktitle}{\emph{Proceedings of the 23rd ACM international
  conference on Multimedia}}. ACM, \bibinfo{pages}{581--590}.
\newblock


\bibitem[\protect\citeauthoryear{Chang, Yu, Yang, and Xing}{Chang
  et~al\mbox{.}}{2016}]%
        {chang2016they}
\bibfield{author}{\bibinfo{person}{Xiaojun Chang}, \bibinfo{person}{Yao-Liang
  Yu}, \bibinfo{person}{Yi Yang}, {and} \bibinfo{person}{Eric~P Xing}.}
  \bibinfo{year}{2016}\natexlab{}.
\newblock \showarticletitle{They are not equally reliable: Semantic event
  search using differentiated concept classifiers}. In
  \bibinfo{booktitle}{\emph{Proceedings of the IEEE Conference on Computer
  Vision and Pattern Recognition}}. \bibinfo{pages}{1884--1893}.
\newblock


\bibitem[\protect\citeauthoryear{Chang, Yu, Yang, and Xing}{Chang
  et~al\mbox{.}}{2017}]%
        {chang2017semantic}
\bibfield{author}{\bibinfo{person}{Xiaojun Chang}, \bibinfo{person}{Yao-Liang
  Yu}, \bibinfo{person}{Yi Yang}, {and} \bibinfo{person}{Eric~P Xing}.}
  \bibinfo{year}{2017}\natexlab{}.
\newblock \showarticletitle{Semantic pooling for complex event analysis in
  untrimmed videos}.
\newblock \bibinfo{journal}{\emph{IEEE transactions on pattern analysis and
  machine intelligence}} \bibinfo{volume}{39}, \bibinfo{number}{8}
  (\bibinfo{year}{2017}), \bibinfo{pages}{1617--1632}.
\newblock


\bibitem[\protect\citeauthoryear{Chen, Cui, Ye, Liu, and Chang}{Chen
  et~al\mbox{.}}{2014}]%
        {chen2014event}
\bibfield{author}{\bibinfo{person}{Jiawei Chen}, \bibinfo{person}{Yin Cui},
  \bibinfo{person}{Guangnan Ye}, \bibinfo{person}{Dong Liu}, {and}
  \bibinfo{person}{Shih-Fu Chang}.} \bibinfo{year}{2014}\natexlab{}.
\newblock \showarticletitle{Event-driven semantic concept discovery by
  exploiting weakly tagged internet images}. In
  \bibinfo{booktitle}{\emph{Proceedings of International Conference on
  Multimedia Retrieval}}. ACM, \bibinfo{pages}{1}.
\newblock


\bibitem[\protect\citeauthoryear{Chen, Dai, Liu, Chen, Yuan, and Liu}{Chen
  et~al\mbox{.}}{2019}]%
        {chen2019dynamic}
\bibfield{author}{\bibinfo{person}{Yinpeng Chen}, \bibinfo{person}{Xiyang Dai},
  \bibinfo{person}{Mengchen Liu}, \bibinfo{person}{Dongdong Chen},
  \bibinfo{person}{Lu Yuan}, {and} \bibinfo{person}{Zicheng Liu}.}
  \bibinfo{year}{2019}\natexlab{}.
\newblock \showarticletitle{Dynamic Convolution: Attention over Convolution
  Kernels}.
\newblock \bibinfo{journal}{\emph{arXiv: Computer Vision and Pattern
  Recognition}} (\bibinfo{year}{2019}).
\newblock


\bibitem[\protect\citeauthoryear{De~Brabandere, Jia, Tuytelaars, and
  Van~Gool}{De~Brabandere et~al\mbox{.}}{2016}]%
        {de-brabandere2016dynamic}
\bibfield{author}{\bibinfo{person}{Bert De~Brabandere}, \bibinfo{person}{Xu
  Jia}, \bibinfo{person}{Tinne Tuytelaars}, {and} \bibinfo{person}{Luc
  Van~Gool}.} \bibinfo{year}{2016}\natexlab{}.
\newblock \showarticletitle{Dynamic filter networks}.
\newblock  (\bibinfo{year}{2016}), \bibinfo{pages}{667--675}.
\newblock


\bibitem[\protect\citeauthoryear{Fan, Chang, Cheng, Yang, Xu, and
  Hauptmann}{Fan et~al\mbox{.}}{2017}]%
        {fan2017complex}
\bibfield{author}{\bibinfo{person}{Hehe Fan}, \bibinfo{person}{Xiaojun Chang},
  \bibinfo{person}{De Cheng}, \bibinfo{person}{Yi Yang}, \bibinfo{person}{Dong
  Xu}, {and} \bibinfo{person}{Alexander~G Hauptmann}.}
  \bibinfo{year}{2017}\natexlab{}.
\newblock \showarticletitle{Complex event detection by identifying reliable
  shots from untrimmed videos}. In \bibinfo{booktitle}{\emph{Proceedings of the
  IEEE International Conference on Computer Vision}}.
  \bibinfo{pages}{736--744}.
\newblock


\bibitem[\protect\citeauthoryear{Feichtenhofer, Fan, Malik, and
  He}{Feichtenhofer et~al\mbox{.}}{2019}]%
        {feichtenhofer2019slowfast}
\bibfield{author}{\bibinfo{person}{Christoph Feichtenhofer},
  \bibinfo{person}{Haoqi Fan}, \bibinfo{person}{Jitendra Malik}, {and}
  \bibinfo{person}{Kaiming He}.} \bibinfo{year}{2019}\natexlab{}.
\newblock \showarticletitle{Slowfast networks for video recognition}. In
  \bibinfo{booktitle}{\emph{Proceedings of the IEEE International Conference on
  Computer Vision}}. \bibinfo{pages}{6202--6211}.
\newblock


\bibitem[\protect\citeauthoryear{Feichtenhofer, Pinz, and
  Zisserman}{Feichtenhofer et~al\mbox{.}}{2016}]%
        {feichtenhofer2016convolutional}
\bibfield{author}{\bibinfo{person}{Christoph Feichtenhofer},
  \bibinfo{person}{Axel Pinz}, {and} \bibinfo{person}{Andrew Zisserman}.}
  \bibinfo{year}{2016}\natexlab{}.
\newblock \showarticletitle{Convolutional two-stream network fusion for video
  action recognition}. In \bibinfo{booktitle}{\emph{Proceedings of the IEEE
  conference on computer vision and pattern recognition}}.
  \bibinfo{pages}{1933--1941}.
\newblock


\bibitem[\protect\citeauthoryear{Gao, Oh, Grauman, and Torresani}{Gao
  et~al\mbox{.}}{2019}]%
        {gao2019listen}
\bibfield{author}{\bibinfo{person}{Ruohan Gao}, \bibinfo{person}{Taehyun Oh},
  \bibinfo{person}{Kristen Grauman}, {and} \bibinfo{person}{Lorenzo
  Torresani}.} \bibinfo{year}{2019}\natexlab{}.
\newblock \showarticletitle{Listen to Look: Action Recognition by Previewing
  Audio}.
\newblock \bibinfo{journal}{\emph{arXiv: Computer Vision and Pattern
  Recognition}} (\bibinfo{year}{2019}).
\newblock


\bibitem[\protect\citeauthoryear{He, Zhang, Ren, and Sun}{He
  et~al\mbox{.}}{2016}]%
        {he2016deep}
\bibfield{author}{\bibinfo{person}{Kaiming He}, \bibinfo{person}{Xiangyu
  Zhang}, \bibinfo{person}{Shaoqing Ren}, {and} \bibinfo{person}{Jian Sun}.}
  \bibinfo{year}{2016}\natexlab{}.
\newblock \showarticletitle{Deep residual learning for image recognition}. In
  \bibinfo{booktitle}{\emph{Proceedings of the IEEE conference on computer
  vision and pattern recognition}}. \bibinfo{pages}{770--778}.
\newblock


\bibitem[\protect\citeauthoryear{Hien}{Hien}{[n.d.]}]%
        {receptive-field-definition}
\bibfield{author}{\bibinfo{person}{Dang Ha~The Hien}.}
  \bibinfo{year}{[n.d.]}\natexlab{}.
\newblock \bibinfo{title}{A Guide to Receptive Field Arithmetic for
  Convolutional Neural Networks}.
\newblock
  \bibinfo{howpublished}{\url{https://syncedreview.com/2017/05/11/a-guide-to-receptive-field-arithmetic-for-convolutional-neural-networks/}}.
\newblock


\bibitem[\protect\citeauthoryear{Jiang, Dai, Wang, Ngo, Xue, and Chang}{Jiang
  et~al\mbox{.}}{2012}]%
        {jiang2012fast}
\bibfield{author}{\bibinfo{person}{Yu-Gang Jiang}, \bibinfo{person}{Qi Dai},
  \bibinfo{person}{Jun Wang}, \bibinfo{person}{Chong-Wah Ngo},
  \bibinfo{person}{Xiangyang Xue}, {and} \bibinfo{person}{Shih-Fu Chang}.}
  \bibinfo{year}{2012}\natexlab{}.
\newblock \showarticletitle{Fast semantic diffusion for large-scale
  context-based image and video annotation}.
\newblock \bibinfo{journal}{\emph{IEEE Transactions on Image Processing}}
  \bibinfo{volume}{21}, \bibinfo{number}{6} (\bibinfo{year}{2012}),
  \bibinfo{pages}{3080--3091}.
\newblock


\bibitem[\protect\citeauthoryear{Jiang, Wu, Wang, Xue, and Chang}{Jiang
  et~al\mbox{.}}{2018}]%
        {jiang2018exploiting}
\bibfield{author}{\bibinfo{person}{Yu-Gang Jiang}, \bibinfo{person}{Zuxuan Wu},
  \bibinfo{person}{Jun Wang}, \bibinfo{person}{Xiangyang Xue}, {and}
  \bibinfo{person}{Shih-Fu Chang}.} \bibinfo{year}{2018}\natexlab{}.
\newblock \showarticletitle{Exploiting feature and class relationships in video
  categorization with regularized deep neural networks}.
\newblock \bibinfo{journal}{\emph{IEEE transactions on pattern analysis and
  machine intelligence}} \bibinfo{volume}{40}, \bibinfo{number}{2}
  (\bibinfo{year}{2018}), \bibinfo{pages}{352--364}.
\newblock


\bibitem[\protect\citeauthoryear{Kang, Kim, Choi, Kim, and Yoo}{Kang
  et~al\mbox{.}}{2018}]%
        {kang2018pivot}
\bibfield{author}{\bibinfo{person}{Sunghun Kang}, \bibinfo{person}{Junyeong
  Kim}, \bibinfo{person}{Hyunsoo Choi}, \bibinfo{person}{Sungjin Kim}, {and}
  \bibinfo{person}{Chang~D Yoo}.} \bibinfo{year}{2018}\natexlab{}.
\newblock \showarticletitle{Pivot Correlational Neural Network for Multimodal
  Video Categorization}. In \bibinfo{booktitle}{\emph{Proceedings of the
  European Conference on Computer Vision (ECCV)}}. \bibinfo{pages}{386--401}.
\newblock


\bibitem[\protect\citeauthoryear{Karpathy, Toderici, Shetty, Leung, Sukthankar,
  and Fei-Fei}{Karpathy et~al\mbox{.}}{2014}]%
        {karpathy2014large}
\bibfield{author}{\bibinfo{person}{Andrej Karpathy}, \bibinfo{person}{George
  Toderici}, \bibinfo{person}{Sanketh Shetty}, \bibinfo{person}{Thomas Leung},
  \bibinfo{person}{Rahul Sukthankar}, {and} \bibinfo{person}{Li Fei-Fei}.}
  \bibinfo{year}{2014}\natexlab{}.
\newblock \showarticletitle{Large-scale video classification with convolutional
  neural networks}. In \bibinfo{booktitle}{\emph{Proceedings of the IEEE
  conference on Computer Vision and Pattern Recognition}}.
  \bibinfo{pages}{1725--1732}.
\newblock


\bibitem[\protect\citeauthoryear{Kay, Carreira, Simonyan, Zhang, Hillier,
  Vijayanarasimhan, Viola, Green, Back, Natsev, et~al\mbox{.}}{Kay
  et~al\mbox{.}}{2017}]%
        {kay2017kinetics}
\bibfield{author}{\bibinfo{person}{Will Kay}, \bibinfo{person}{Joao Carreira},
  \bibinfo{person}{Karen Simonyan}, \bibinfo{person}{Brian Zhang},
  \bibinfo{person}{Chloe Hillier}, \bibinfo{person}{Sudheendra
  Vijayanarasimhan}, \bibinfo{person}{Fabio Viola}, \bibinfo{person}{Tim
  Green}, \bibinfo{person}{Trevor Back}, \bibinfo{person}{Paul Natsev},
  {et~al\mbox{.}}} \bibinfo{year}{2017}\natexlab{}.
\newblock \showarticletitle{The kinetics human action video dataset}.
\newblock \bibinfo{journal}{\emph{arXiv preprint arXiv:1705.06950}}
  (\bibinfo{year}{2017}).
\newblock


\bibitem[\protect\citeauthoryear{Kloft, Brefeld, Sonnenburg, and Zien}{Kloft
  et~al\mbox{.}}{2011}]%
        {kloft2011lp}
\bibfield{author}{\bibinfo{person}{Marius Kloft}, \bibinfo{person}{Ulf
  Brefeld}, \bibinfo{person}{S{\"o}ren Sonnenburg}, {and}
  \bibinfo{person}{Alexander Zien}.} \bibinfo{year}{2011}\natexlab{}.
\newblock \showarticletitle{Lp-norm multiple kernel learning}.
\newblock \bibinfo{journal}{\emph{Journal of Machine Learning Research}}
  \bibinfo{volume}{12}, \bibinfo{number}{Mar} (\bibinfo{year}{2011}),
  \bibinfo{pages}{953--997}.
\newblock


\bibitem[\protect\citeauthoryear{Li, Wang, Hu, and Yang}{Li
  et~al\mbox{.}}{2019b}]%
        {li2019selective}
\bibfield{author}{\bibinfo{person}{Xiang Li}, \bibinfo{person}{Wenhai Wang},
  \bibinfo{person}{Xiaolin Hu}, {and} \bibinfo{person}{Jian Yang}.}
  \bibinfo{year}{2019}\natexlab{b}.
\newblock \showarticletitle{Selective Kernel Networks}.
\newblock  (\bibinfo{year}{2019}), \bibinfo{pages}{510--519}.
\newblock


\bibitem[\protect\citeauthoryear{Li, Chen, Wang, and Zhang}{Li
  et~al\mbox{.}}{2019a}]%
        {li2019scale-aware}
\bibfield{author}{\bibinfo{person}{Yanghao Li}, \bibinfo{person}{Yuntao Chen},
  \bibinfo{person}{Naiyan Wang}, {and} \bibinfo{person}{Zhaoxiang Zhang}.}
  \bibinfo{year}{2019}\natexlab{a}.
\newblock \showarticletitle{Scale-Aware Trident Networks for Object Detection}.
\newblock \bibinfo{journal}{\emph{arXiv: Computer Vision and Pattern
  Recognition}} (\bibinfo{year}{2019}).
\newblock


\bibitem[\protect\citeauthoryear{Lioutas and Guo}{Lioutas and Guo}{2020}]%
        {lioutas2020time-aware}
\bibfield{author}{\bibinfo{person}{Vasileios Lioutas} {and}
  \bibinfo{person}{Yuhong Guo}.} \bibinfo{year}{2020}\natexlab{}.
\newblock \showarticletitle{Time-aware Large Kernel Convolutions}.
\newblock \bibinfo{journal}{\emph{arXiv: Learning}} (\bibinfo{year}{2020}).
\newblock


\bibitem[\protect\citeauthoryear{Long, Gan, De~Melo, Liu, Li, Li, and Wen}{Long
  et~al\mbox{.}}{2018}]%
        {long2018multimodal}
\bibfield{author}{\bibinfo{person}{Xiang Long}, \bibinfo{person}{Chuang Gan},
  \bibinfo{person}{Gerard De~Melo}, \bibinfo{person}{Xiao Liu},
  \bibinfo{person}{Yandong Li}, \bibinfo{person}{Fu Li}, {and}
  \bibinfo{person}{Shilei Wen}.} \bibinfo{year}{2018}\natexlab{}.
\newblock \showarticletitle{Multimodal keyless attention fusion for video
  classification}. In \bibinfo{booktitle}{\emph{Thirty-Second AAAI Conference
  on Artificial Intelligence}}.
\newblock


\bibitem[\protect\citeauthoryear{Qiu, Yao, and Mei}{Qiu et~al\mbox{.}}{2017}]%
        {qiu2017learning}
\bibfield{author}{\bibinfo{person}{Zhaofan Qiu}, \bibinfo{person}{Ting Yao},
  {and} \bibinfo{person}{Tao Mei}.} \bibinfo{year}{2017}\natexlab{}.
\newblock \showarticletitle{Learning spatio-temporal representation with
  pseudo-3d residual networks}. In \bibinfo{booktitle}{\emph{proceedings of the
  IEEE International Conference on Computer Vision}}.
  \bibinfo{pages}{5533--5541}.
\newblock


\bibitem[\protect\citeauthoryear{Russakovsky, Deng, Su, Krause, Satheesh, Ma,
  Huang, Karpathy, Khosla, Bernstein, et~al\mbox{.}}{Russakovsky
  et~al\mbox{.}}{2015}]%
        {russakovsky2015imagenet}
\bibfield{author}{\bibinfo{person}{Olga Russakovsky}, \bibinfo{person}{Jia
  Deng}, \bibinfo{person}{Hao Su}, \bibinfo{person}{Jonathan Krause},
  \bibinfo{person}{Sanjeev Satheesh}, \bibinfo{person}{Sean Ma},
  \bibinfo{person}{Zhiheng Huang}, \bibinfo{person}{Andrej Karpathy},
  \bibinfo{person}{Aditya Khosla}, \bibinfo{person}{Michael Bernstein},
  {et~al\mbox{.}}} \bibinfo{year}{2015}\natexlab{}.
\newblock \showarticletitle{Imagenet large scale visual recognition challenge}.
\newblock \bibinfo{journal}{\emph{International journal of computer vision}}
  \bibinfo{volume}{115}, \bibinfo{number}{3} (\bibinfo{year}{2015}),
  \bibinfo{pages}{211--252}.
\newblock


\bibitem[\protect\citeauthoryear{Smith, Naphade, and Natsev}{Smith
  et~al\mbox{.}}{2003}]%
        {smith2003multimedia}
\bibfield{author}{\bibinfo{person}{John~R Smith}, \bibinfo{person}{Milind
  Naphade}, {and} \bibinfo{person}{Apostol Natsev}.}
  \bibinfo{year}{2003}\natexlab{}.
\newblock \showarticletitle{Multimedia semantic indexing using model vectors}.
  In \bibinfo{booktitle}{\emph{2003 International Conference on Multimedia and
  Expo. ICME'03. Proceedings (Cat. No. 03TH8698)}}, Vol.~\bibinfo{volume}{2}.
  IEEE, \bibinfo{pages}{II--445}.
\newblock


\bibitem[\protect\citeauthoryear{Srivastava and Salakhutdinov}{Srivastava and
  Salakhutdinov}{2012}]%
        {srivastava2012multimodal}
\bibfield{author}{\bibinfo{person}{Nitish Srivastava} {and}
  \bibinfo{person}{Ruslan~R Salakhutdinov}.} \bibinfo{year}{2012}\natexlab{}.
\newblock \showarticletitle{Multimodal learning with deep boltzmann machines}.
  In \bibinfo{booktitle}{\emph{Advances in neural information processing
  systems}}. \bibinfo{pages}{2222--2230}.
\newblock


\bibitem[\protect\citeauthoryear{Szegedy, Ioffe, Vanhoucke, and Alemi}{Szegedy
  et~al\mbox{.}}{2016a}]%
        {szegedy2016inception-v4}
\bibfield{author}{\bibinfo{person}{Christian Szegedy}, \bibinfo{person}{Sergey
  Ioffe}, \bibinfo{person}{Vincent Vanhoucke}, {and}
  \bibinfo{person}{Alexander~A Alemi}.} \bibinfo{year}{2016}\natexlab{a}.
\newblock \showarticletitle{Inception-v4, Inception-ResNet and the Impact of
  Residual Connections on Learning}.
\newblock  (\bibinfo{year}{2016}), \bibinfo{pages}{4278--4284}.
\newblock


\bibitem[\protect\citeauthoryear{Szegedy, Vanhoucke, Ioffe, Shlens, and
  Wojna}{Szegedy et~al\mbox{.}}{2016b}]%
        {szegedy2016rethinking}
\bibfield{author}{\bibinfo{person}{Christian Szegedy}, \bibinfo{person}{Vincent
  Vanhoucke}, \bibinfo{person}{Sergey Ioffe}, \bibinfo{person}{Jonathon
  Shlens}, {and} \bibinfo{person}{Zbigniew Wojna}.}
  \bibinfo{year}{2016}\natexlab{b}.
\newblock \showarticletitle{Rethinking the Inception Architecture for Computer
  Vision}.
\newblock  (\bibinfo{year}{2016}), \bibinfo{pages}{2818--2826}.
\newblock


\bibitem[\protect\citeauthoryear{Tran, Bourdev, Fergus, Torresani, and
  Paluri}{Tran et~al\mbox{.}}{2015}]%
        {tran2015learning}
\bibfield{author}{\bibinfo{person}{Du Tran}, \bibinfo{person}{Lubomir Bourdev},
  \bibinfo{person}{Rob Fergus}, \bibinfo{person}{Lorenzo Torresani}, {and}
  \bibinfo{person}{Manohar Paluri}.} \bibinfo{year}{2015}\natexlab{}.
\newblock \showarticletitle{Learning spatiotemporal features with 3d
  convolutional networks}. In \bibinfo{booktitle}{\emph{Proceedings of the IEEE
  international conference on computer vision}}. \bibinfo{pages}{4489--4497}.
\newblock


\bibitem[\protect\citeauthoryear{Wang, Xiong, Wang, Qiao, Lin, Tang, and
  Van~Gool}{Wang et~al\mbox{.}}{2016}]%
        {wang2016temporal}
\bibfield{author}{\bibinfo{person}{Limin Wang}, \bibinfo{person}{Yuanjun
  Xiong}, \bibinfo{person}{Zhe Wang}, \bibinfo{person}{Yu Qiao},
  \bibinfo{person}{Dahua Lin}, \bibinfo{person}{Xiaoou Tang}, {and}
  \bibinfo{person}{Luc Van~Gool}.} \bibinfo{year}{2016}\natexlab{}.
\newblock \showarticletitle{Temporal segment networks: Towards good practices
  for deep action recognition}. In \bibinfo{booktitle}{\emph{European
  conference on computer vision}}. Springer, \bibinfo{pages}{20--36}.
\newblock


\bibitem[\protect\citeauthoryear{Wu, He, Tan, Chen, and Wen}{Wu
  et~al\mbox{.}}{2019a}]%
        {wu2019multi}
\bibfield{author}{\bibinfo{person}{Wenhao Wu}, \bibinfo{person}{Dongliang He},
  \bibinfo{person}{Xiao Tan}, \bibinfo{person}{Shifeng Chen}, {and}
  \bibinfo{person}{Shilei Wen}.} \bibinfo{year}{2019}\natexlab{a}.
\newblock \showarticletitle{Multi-Agent Reinforcement Learning Based Frame
  Sampling for Effective Untrimmed Video Recognition}. In
  \bibinfo{booktitle}{\emph{Proceedings of the IEEE International Conference on
  Computer Vision}}. \bibinfo{pages}{6222--6231}.
\newblock


\bibitem[\protect\citeauthoryear{Wu, Fu, Jiang, and Sigal}{Wu
  et~al\mbox{.}}{2016}]%
        {wu2016harnessing}
\bibfield{author}{\bibinfo{person}{Zuxuan Wu}, \bibinfo{person}{Yanwei Fu},
  \bibinfo{person}{Yu-Gang Jiang}, {and} \bibinfo{person}{Leonid Sigal}.}
  \bibinfo{year}{2016}\natexlab{}.
\newblock \showarticletitle{Harnessing object and scene semantics for
  large-scale video understanding}. In \bibinfo{booktitle}{\emph{Proceedings of
  the IEEE Conference on Computer Vision and Pattern Recognition}}.
  \bibinfo{pages}{3112--3121}.
\newblock


\bibitem[\protect\citeauthoryear{Wu, Xiong, Jiang, and Davis}{Wu
  et~al\mbox{.}}{2019b}]%
        {wu2019liteeval}
\bibfield{author}{\bibinfo{person}{Zuxuan Wu}, \bibinfo{person}{Caiming Xiong},
  \bibinfo{person}{Yu-Gang Jiang}, {and} \bibinfo{person}{Larry~S Davis}.}
  \bibinfo{year}{2019}\natexlab{b}.
\newblock \showarticletitle{LiteEval: A Coarse-to-Fine Framework for Resource
  Efficient Video Recognition}. In \bibinfo{booktitle}{\emph{Advances in Neural
  Information Processing Systems}}. \bibinfo{pages}{7778--7787}.
\newblock


\bibitem[\protect\citeauthoryear{Wu, Xiong, Ma, Socher, and Davis}{Wu
  et~al\mbox{.}}{2019c}]%
        {wu2019adaframe}
\bibfield{author}{\bibinfo{person}{Zuxuan Wu}, \bibinfo{person}{Caiming Xiong},
  \bibinfo{person}{Chih-Yao Ma}, \bibinfo{person}{Richard Socher}, {and}
  \bibinfo{person}{Larry~S Davis}.} \bibinfo{year}{2019}\natexlab{c}.
\newblock \showarticletitle{AdaFrame: Adaptive Frame Selection for Fast Video
  Recognition}. In \bibinfo{booktitle}{\emph{Proceedings of the IEEE Conference
  on Computer Vision and Pattern Recognition}}. \bibinfo{pages}{1278--1287}.
\newblock


\bibitem[\protect\citeauthoryear{Xu, Tsang, Yang, Ma, and Hauptmann}{Xu
  et~al\mbox{.}}{2014}]%
        {xu2014event}
\bibfield{author}{\bibinfo{person}{Zhongwen Xu}, \bibinfo{person}{Ivor~W
  Tsang}, \bibinfo{person}{Yi Yang}, \bibinfo{person}{Zhigang Ma}, {and}
  \bibinfo{person}{Alexander~G Hauptmann}.} \bibinfo{year}{2014}\natexlab{}.
\newblock \showarticletitle{Event detection using multi-level relevance labels
  and multiple features}. In \bibinfo{booktitle}{\emph{Proceedings of the IEEE
  Conference on Computer Vision and Pattern Recognition}}.
  \bibinfo{pages}{97--104}.
\newblock


\bibitem[\protect\citeauthoryear{Yan, Yang, Shen, Meng, Liu, Hauptmann, and
  Sebe}{Yan et~al\mbox{.}}{2015}]%
        {yan2015complex}
\bibfield{author}{\bibinfo{person}{Yan Yan}, \bibinfo{person}{Yi Yang},
  \bibinfo{person}{Haoquan Shen}, \bibinfo{person}{Deyu Meng},
  \bibinfo{person}{Gaowen Liu}, \bibinfo{person}{Alex Hauptmann}, {and}
  \bibinfo{person}{Nicu Sebe}.} \bibinfo{year}{2015}\natexlab{}.
\newblock \showarticletitle{Complex event detection via event oriented
  dictionary learning}. In \bibinfo{booktitle}{\emph{Twenty-Ninth AAAI
  Conference on Artificial Intelligence}}.
\newblock


\bibitem[\protect\citeauthoryear{Yang, Bender, Le, and Ngiam}{Yang
  et~al\mbox{.}}{2019}]%
        {yang2019condconv:}
\bibfield{author}{\bibinfo{person}{Brandon Yang}, \bibinfo{person}{Gabriel
  Bender}, \bibinfo{person}{Quoc~V Le}, {and} \bibinfo{person}{Jiquan Ngiam}.}
  \bibinfo{year}{2019}\natexlab{}.
\newblock \showarticletitle{CondConv: Conditionally Parameterized Convolutions
  for Efficient Inference}.
\newblock \bibinfo{journal}{\emph{arXiv: Computer Vision and Pattern
  Recognition}} (\bibinfo{year}{2019}).
\newblock


\bibitem[\protect\citeauthoryear{Ye, Li, Xu, Liu, and Chang}{Ye
  et~al\mbox{.}}{2015}]%
        {ye2015eventnet}
\bibfield{author}{\bibinfo{person}{Guangnan Ye}, \bibinfo{person}{Yitong Li},
  \bibinfo{person}{Hongliang Xu}, \bibinfo{person}{Dong Liu}, {and}
  \bibinfo{person}{Shih-Fu Chang}.} \bibinfo{year}{2015}\natexlab{}.
\newblock \showarticletitle{Eventnet: A large scale structured concept library
  for complex event detection in video}. In
  \bibinfo{booktitle}{\emph{Proceedings of the 23rd ACM international
  conference on Multimedia}}. ACM, \bibinfo{pages}{471--480}.
\newblock


\bibitem[\protect\citeauthoryear{Yu and Koltun}{Yu and Koltun}{2016}]%
        {Yu2016Multi}
\bibfield{author}{\bibinfo{person}{Fisher Yu} {and} \bibinfo{person}{Vladlen
  Koltun}.} \bibinfo{year}{2016}\natexlab{}.
\newblock \showarticletitle{Multi-Scale Context Aggregation by Dilated
  Convolutions}.
\newblock  (\bibinfo{year}{2016}).
\newblock


\bibitem[\protect\citeauthoryear{Zhang, Mei, Zheng, and Fan}{Zhang
  et~al\mbox{.}}{2019b}]%
        {zhang2019exploiting}
\bibfield{author}{\bibinfo{person}{Ji Zhang}, \bibinfo{person}{Kuizhi Mei},
  \bibinfo{person}{Yu Zheng}, {and} \bibinfo{person}{Jianping Fan}.}
  \bibinfo{year}{2019}\natexlab{b}.
\newblock \showarticletitle{Exploiting Mid-Level Semantics for Large-Scale
  Complex Video Classification}.
\newblock \bibinfo{journal}{\emph{IEEE Transactions on Multimedia}}
  (\bibinfo{year}{2019}).
\newblock


\bibitem[\protect\citeauthoryear{Zhang, Halber, and Rusinkiewicz}{Zhang
  et~al\mbox{.}}{2019a}]%
        {zhang2019accelerating}
\bibfield{author}{\bibinfo{person}{Linguang Zhang}, \bibinfo{person}{Maciej
  Halber}, {and} \bibinfo{person}{Szymon Rusinkiewicz}.}
  \bibinfo{year}{2019}\natexlab{a}.
\newblock \showarticletitle{Accelerating Large-Kernel Convolution Using
  Summed-Area Tables.}
\newblock \bibinfo{journal}{\emph{arXiv: Learning}} (\bibinfo{year}{2019}).
\newblock


\bibitem[\protect\citeauthoryear{Zhao, Zhang, Wu, Li, and Jiang}{Zhao
  et~al\mbox{.}}{2019}]%
        {zhao2019visual}
\bibfield{author}{\bibinfo{person}{Rui-Wei Zhao}, \bibinfo{person}{Qi Zhang},
  \bibinfo{person}{Zuxuan Wu}, \bibinfo{person}{Jianguo Li}, {and}
  \bibinfo{person}{Yu-Gang Jiang}.} \bibinfo{year}{2019}\natexlab{}.
\newblock \showarticletitle{Visual Content Recognition by Exploiting Semantic
  Feature Map with Attention and Multi-task Learning}.
\newblock \bibinfo{journal}{\emph{ACM Transactions on Multimedia Computing,
  Communications, and Applications (TOMM)}} \bibinfo{volume}{15},
  \bibinfo{number}{1s} (\bibinfo{year}{2019}), \bibinfo{pages}{6}.
\newblock


\bibitem[\protect\citeauthoryear{Zhou, Lapedriza, Khosla, Oliva, and
  Torralba}{Zhou et~al\mbox{.}}{2018}]%
        {zhou2018places}
\bibfield{author}{\bibinfo{person}{Bolei Zhou}, \bibinfo{person}{Agata
  Lapedriza}, \bibinfo{person}{Aditya Khosla}, \bibinfo{person}{Aude Oliva},
  {and} \bibinfo{person}{Antonio Torralba}.} \bibinfo{year}{2018}\natexlab{}.
\newblock \showarticletitle{Places: A 10 million image database for scene
  recognition}.
\newblock \bibinfo{journal}{\emph{IEEE transactions on pattern analysis and
  machine intelligence}} \bibinfo{volume}{40}, \bibinfo{number}{6}
  (\bibinfo{year}{2018}), \bibinfo{pages}{1452--1464}.
\newblock


\end{thebibliography}

\appendix

\end{document}